\documentclass[10pt,twocolumn,letterpaper]{article}

\usepackage[pagenumbers]{cvpr} 

\usepackage{graphicx}
\usepackage{adjustbox}
\usepackage{kotex}
\usepackage{colortbl}
\usepackage{subcaption}
\usepackage{xcolor}

\usepackage{changepage} 
\usepackage{threeparttable} 

\newcommand{\method}{DefectFill}
\newcommand{\ldm}{LDM}

\definecolor{cvprblue}{rgb}{0.21,0.49,0.74}
\usepackage[pagebackref,breaklinks,colorlinks,allcolors=cvprblue]{hyperref}

\usepackage{amsmath}
\usepackage{kotex}
\usepackage{graphicx}
\usepackage{multirow}
\usepackage{caption}
\usepackage{subcaption}
\usepackage{booktabs}
\usepackage{comment}
\usepackage{xcolor}
\usepackage{fancyhdr} 

\usepackage[accsupp]{axessibility}  

\def\paperID{15135} 
\def\confName{CVPR}
\def\confYear{2025}

\title{DefectFill: Realistic Defect Generation with Inpainting Diffusion Model for Visual Inspection}

\author{Jaewoo Song$^{1,2}$~~~~~~~~~~Daemin Park$^{1}$~~~~~~~~~~Kanghyun Baek$^{3}$~~~~~~~~~~Sangyub Lee$^{3}$\\
Jooyoung Choi$^{1}$~~~~~~~~~~Eunji Kim$^{1}$~~~~~~~~~~Sungroh Yoon$^{1,3,4,}$\footnotemark[1]\\
$^1$Department of Electrical and Computer Engineering, Seoul National University\\
$^2$Global Technology Research, Samsung Electronics\\
$^3$IPAI, $^4$AIIS, ASRI, INMC, ISRC, Seoul National University\\
{\tt\small \{woo.song, eoalsqkr12, qor6271, nickyub, jy\_choi, kce407, sryoon\}@snu.ac.kr}
}

\begin{document}

\twocolumn[{
    \renewcommand\twocolumn[1][]{#1}
    \maketitle
    \begin{center}
    \centering
        \includegraphics[width=\textwidth]{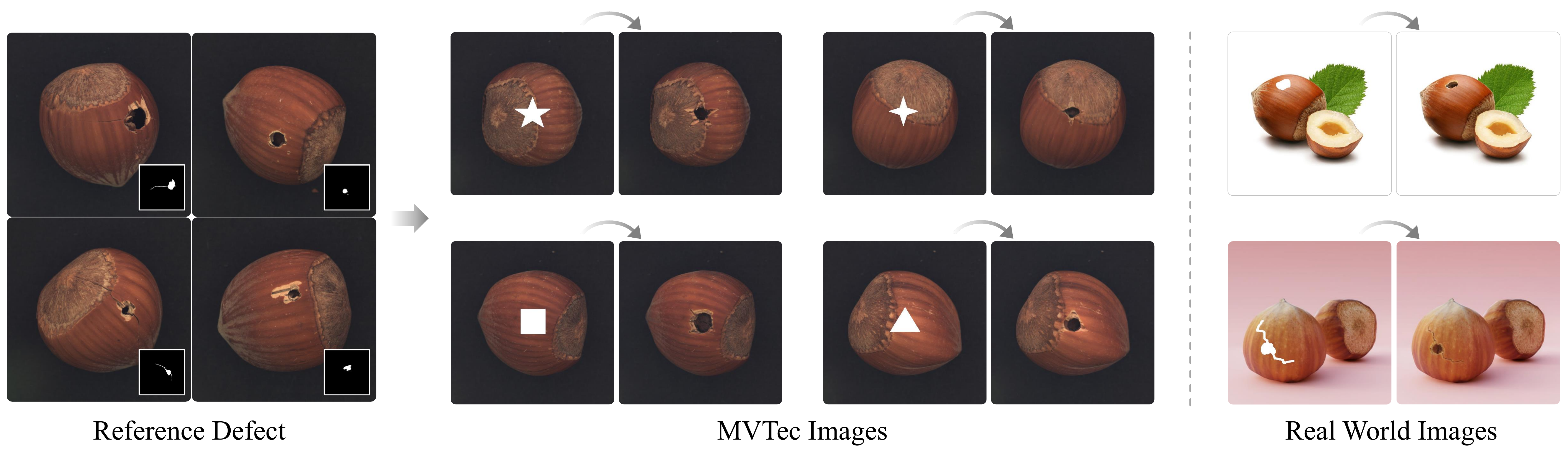}
        \captionof{figure}{Given a few reference image-mask pairs of a defect (\eg ``hole" of a hazelnut), DefectFill learns the defect and realistically fill it onto defect-free objects in desired shapes (\eg star, square, \textit{etc.}), generating new defect images that integrate naturally with the objects. These generated images are then used for visual inspection tasks.}
        \label{fig:main}
    \end{center}
}]

\renewcommand{\thefootnote}{\fnsymbol{footnote}}
\footnotetext[1]{Correspondence to: Sungroh Yoon (sryoon@snu.ac.kr)}

\begin{abstract}

Developing effective visual inspection models remains challenging due to the scarcity of defect data. While image generation models have been used to synthesize defect images, producing highly realistic defects remains difficult. We propose DefectFill, a novel method for realistic defect generation that requires only a few reference defect images. It leverages a fine-tuned inpainting diffusion model, optimized with our custom loss functions incorporating defect, object, and attention terms. It enables precise capture of detailed, localized defect features and their seamless integration into defect-free objects. Additionally, our Low-Fidelity Selection method further enhances the defect sample quality. Experiments show that DefectFill generates high-quality defect images, enabling visual inspection models to achieve state-of-the-art performance on the MVTec AD dataset.

\end{abstract}

\section{Introduction}
\label{sec:intro}

Automating inspection on manufacturing lines is a crucial step in advancing smart factories. In this context, visual inspection focused on defect detection is a critical application for AI models. With substantial amounts of defective data, high-performance models can be developed through supervised learning~\cite{du2021automated}. However, collecting large quantities of defective data is challenging in real-world settings. For example, in newly established production lines or semiconductor processes with exceptionally low defect rates, it may be difficult or even impossible to acquire enough data.

To overcome the limited availability of defect data, various approaches have been developed, including out-of-distribution (OOD) techniques~\cite{ndiour2022subspace} and anomaly detection (AD)~\cite{roth2022towards}, which only use non-defective data, as well as active learning~\cite{lv2020deep} and semi-supervised learning~\cite{ruff2019deep} with limited defective data.
However, these methods have limitations: defect criteria vary across different problems and often require domain expertise, and they struggle to classify defect types accurately. To address these issues, some methods propose generating defect images to train visual inspection models~\cite{zhang2021defect,niu2020defect,duan2023few,hu2024anomalydiffusion}. Yet, a key problem remains: defect images generated by existing methods appear unrealistic, lacking the clarity and natural details of real-world defects, which limits their practical effectiveness.

In this paper, we focus on generating realistic defect images to improve the accuracy of visual inspection tasks. To achieve this, we address two key considerations: (1) precisely capturing defect details and (2) seamlessly incorporating these defect features into defect-free images.

We introduce \method{}, a novel approach for generating realistic and detailed defect images using abundant normal images along with a few reference defect samples. We leverage a pre-trained inpainting diffusion model~\cite{rombach2022high} to remove certain areas of a defect-free image and naturally fill those areas with defects. However, accurately filling defects is challenging, as these features often have entirely different textures or appearances compared to the original object. Therefore, we introduce three loss functions: \textit{defect loss} to capture detailed features of the defect itself, \textit{object loss} to establish the semantic relationship between the defect and the object, and \textit{attention loss} to ensure the word representing the defect focuses precisely on the defect area. These carefully designed loss functions are essential for generating realistic defect images, enabling defects to be naturally and authentically ``filled" within objects. To further refine samples, we implement the Low-Fidelity Selection method, which filters out generated images that fail to represent defects clearly, ensuring only high-quality samples are used.

Through extensive experiments, we demonstrate our model's ability to generate realistic defect images that outperform state-of-the-art methods in both qualitative and quantitative evaluations.
Finally, by leveraging our high-quality generated defect images, we improve performance in visual inspection downstream tasks such as anomaly classification and localization, showing that DefectFill effectively addresses the shortage of defect data.

Our main contributions include: (1) pioneering the use of an inpainting diffusion model for generating defect images, (2) designing novel loss functions that enable the model to learn embedded defect characteristics within the context of the object, thereby generating realistic defects, (3) introducing the Low-Fidelity Selection method which is used to further enhance the quality of generated samples, and (4) demonstrating that our realistic defect images significantly improve performance in downstream tasks.
\begin{figure*}
  \centering
    \includegraphics[width=\textwidth]{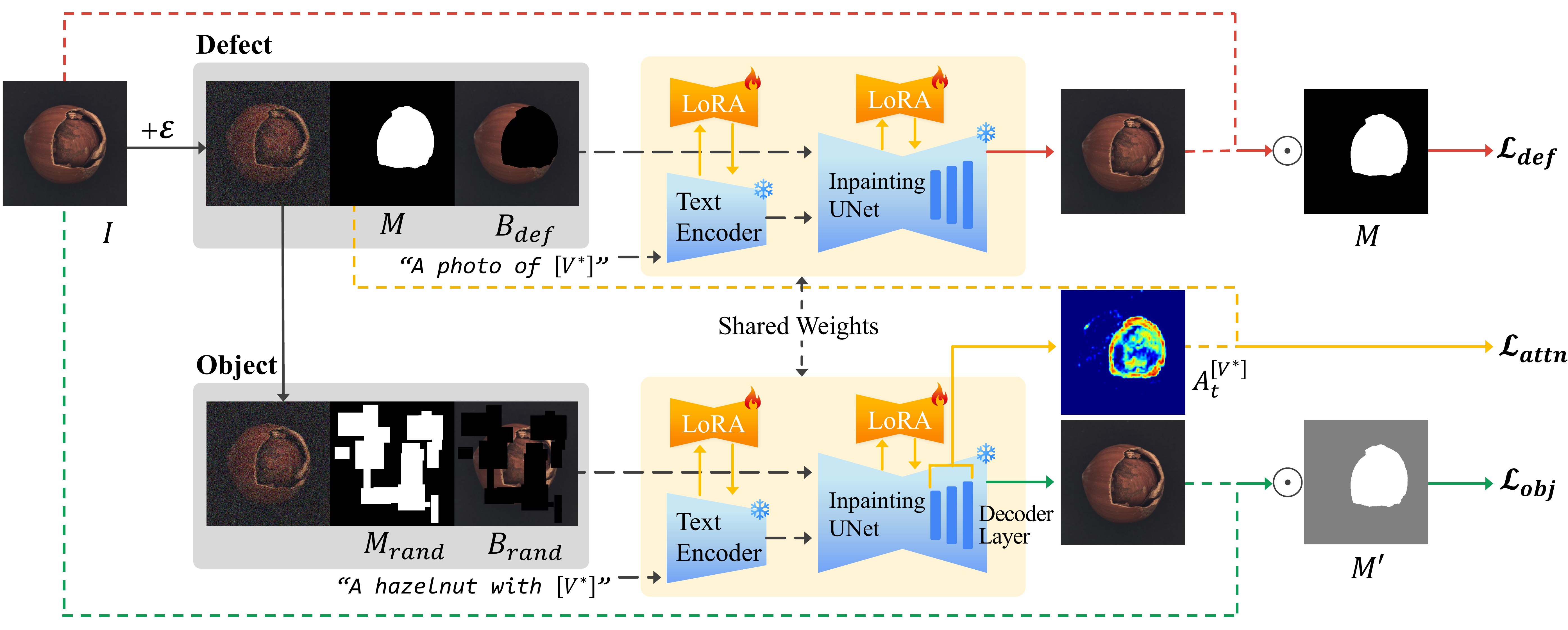}
    \caption{\textbf{Defect learning overview.} To fine-tune the inpainting diffusion model, we compute three types of loss ($\mathcal{L}_{def}$, $\mathcal{L}_{attn}$, and $\mathcal{L}_{obj}$) using an image $I$ and a defect mask $M$. The image $I$ is duplicated, with each copy combined with different masks ($M$ and $M_{rand}$) and prompts ($\mathcal{P}_{def}$: \textit{``A photo of [$V^*$]"}, and $\mathcal{P}_{obj}$: \textit{``A hazelnut with [$V^*$]"}) as inputs to the model. The model prediction using the defect prompt $\mathcal{P}_{def}$ (upper pipeline) is used to compute $\mathcal{L}_{def}$ and, while the prediction using the object prompt $\mathcal{P}_{obj}$ (lower pipeline) is used to compute $\mathcal{L}_{attn}$ and $\mathcal{L}_{obj}$.}
    \label{fig:method overview}
  \hfill
\end{figure*}

\section{Related Work}
\label{sec:related_work}

\subsection{Anomaly generation}
Various approaches have been proposed to mitigate the scarcity of defective data by generating synthetic defects~\cite{lin2021few, zhang2023prototypical, li2021cutpaste, zavrtanik2021draem}.
Crop-Paste~\cite{lin2021few} and CutPaste~\cite{li2021cutpaste} synthesize data by extracting in-distribution image patches and repositioning them, while PRN~\cite{zhang2023prototypical} and DRAEM~\cite{zavrtanik2021draem} incorporate out-of-distribution elements into normal images to generate additional synthetic anomalies.
Since these methods solely rely on data augmentation, their ability to generate truly novel defects remains limited, thus constraining diversity.
Additionally, the defects synthesized using cross-distribution images often lack realism.

Recent research has shifted toward direct defect image generation using Generative Adversarial Networks (GANs)~\cite{goodfellow2020generative}, including methods like SDGAN~\cite{niu2020defect} and Defect-GAN~\cite{zhang2021defect}. However, these approaches require large and diverse defect datasets, which limits their applicability in data-scarce scenarios. DFMGAN~\cite{duan2023few} addresses this limitation by enabling defect image generation from a small number of reference images, by exploiting a pre-trained StyleGAN2~\cite{karras2020analyzingimprovingimagequality}. Nonetheless, it demands lengthy training times and struggles with generating realistic defects. In contrast to GAN-based models, studies using powerful text-to-image diffusion models~\cite{rombach2022high} have shown promising results. AnomalyDiffusion~\cite{hu2024anomalydiffusion} optimizes word vectors to disentangle the intrinsic characteristics of defects from their positional information, allowing defects to be generated at any specified location. However, these word vectors still fall short in capturing the fine structural details of defects~\cite{gal2022image}, resulting in defects that lack realism.

\subsection{Personalization}
Leveraging the text-to-image capabilities of diffusion models, personalization research has emerged to learn new objects unknown to these models. This learning process uses a few reference images to enable a unique word token [$V^*$] to represent the new target concept. Once the concept is learned, prompts containing the [$V^*$] token can be used to generate new images of this concept. Most studies primarily focus on learning a main object that occupies most of the image, either by optimizing the unique word token~\cite{gal2022image} or fine-tuning the diffusion model~\cite{ruiz2023dreambooth, kumari2023multi}.

In contrast, CLiC~\cite{safaee2024clic} focuses on learning local concepts rather than the main object and employs cross-attention guidance~\cite{chefer2023attend} to transfer these local concepts. We draw inspiration from this approach, though it is primarily designed for realistic scenarios where the target object can naturally exhibit these concepts, unlike defect images. In addition to the previously mentioned studies, there has been an effort to use inpainting diffusion models to learn concepts~\cite{tang2024realfill}. This approach focuses on learning a single target image alongside reference images, solely for inpainting that target.

Related to these studies, we aim to learn a defect concept anomalous to objects and generate diverse, realistic defect images to enhance the performance of downstream tasks.
\section{Methods}
\label{sec:methods}

We introduce DefectFill, a novel method for generating diverse and realistic defect images. By fine-tuning a pre-trained inpainting diffusion model, DefectFill efficiently learns defect concepts using only a limited set of reference defect image-mask pairs. During inference, it fills the defect feature into specific areas of defect-free images, thereby enabling the generation of high-quality defect images that enhance performance in visual inspection tasks.

The following sections cover the background on inpainting diffusion models (Sec.\ref{preliminaries}), followed by a formal description of our method for learning defects (Sec.\ref{learn_defect}) and generating defect images with the learned defects (Sec.\ref{generate_defect}). Subsequently, we describe how the generated images can be applied to downstream tasks (Sec.\ref{applying_downstream}).

\subsection{Preliminaries}
\label{preliminaries} 

\paragraph{Latent Diffusion Models.}  
Latent Diffusion Models (\ldm{}s)~\cite{rombach2022high} are a class of diffusion models~\cite{ho2020denoising,songscore,sohl2015deep} specifically designed to enhance efficiency by reducing computational complexity.
An \ldm{} consists of an encoder $\mathcal{E}$ that maps image $I$ to a latent space $x_0 = \mathcal{E}(I)$, a decoder $\mathcal{D}$ that reconstructs images as $I = \mathcal{D}(x_0)$, and a diffusion model operating in the latent space.
The encoder and decoder are pre-trained to accurately reconstruct images from their latent representations such that $\mathcal{D}(\mathcal{E}(I)) = I$, while the diffusion model is trained to predict the noise that needs to be removed from a noisy latent representation.

The forward process of the diffusion model gradually adds Gaussian noise $\epsilon\sim\mathcal{N}(0, 1)$ to the latent image:
\begin{equation}
    x_t = \sqrt{\alpha_t} x_0 + \left(\sqrt{1 - \alpha_t}\right) \epsilon,
    \label{eq:x_t}
\end{equation}
where $\{\alpha_t\}_{t=1}^{T}$ is a noise scheduler that determines the proportion of noise added at each timestep $t$. The reverse process reconstructs the latent image from the noisy input $x_t$. The diffusion model can incorporate a text prompt $\mathcal{P}$ as conditioning, which is encoded by a text encoder $\tau_\theta$, and is trained using the following objective:
\begin{equation}
    \mathcal{L} = \mathbb{E}_{x_t, t, \epsilon} \left\| \epsilon_\theta (x_t, t, \tau_\theta(\mathcal{P})) - \epsilon \right\|_2^2.
    \label{eq:ldm_training}
\end{equation}

\paragraph{Inpainting Diffusion Models.}

Inpainting Diffusion Models are fine-tuned versions of \ldm{}, specifically designed to fill content within masked areas. These models learn the inpainting task using both a mask $M$ and a background image $B$ where the masked areas are removed. Specifically, the image $I$ and the background $B$ are mapped into the latent space through the encoder, resulting in $x_0=\mathcal{E}(I)$ and $b=\mathcal{E}(B)$. Gaussian noise is then added to $x_0$ at a specific timestep $t$, producing $x_t$. Subsequently, $x_{t}^{inpaint}$ is constructed by concatenating $x_t$, $b$, and $M$ as follows:
\begin{equation}
x_{t}^{inpaint}=\mathrm{concat}\left(x_t,b,M\right)
  \label{eq:x_inpaint},
\end{equation}
and is used for training diffusion models via \cref{eq:ldm_training}. 
This process allows the \ldm{} to learn how to accurately fill the masked areas with appropriate content.

\begin{figure}
  \centering
  \includegraphics[width=\columnwidth]{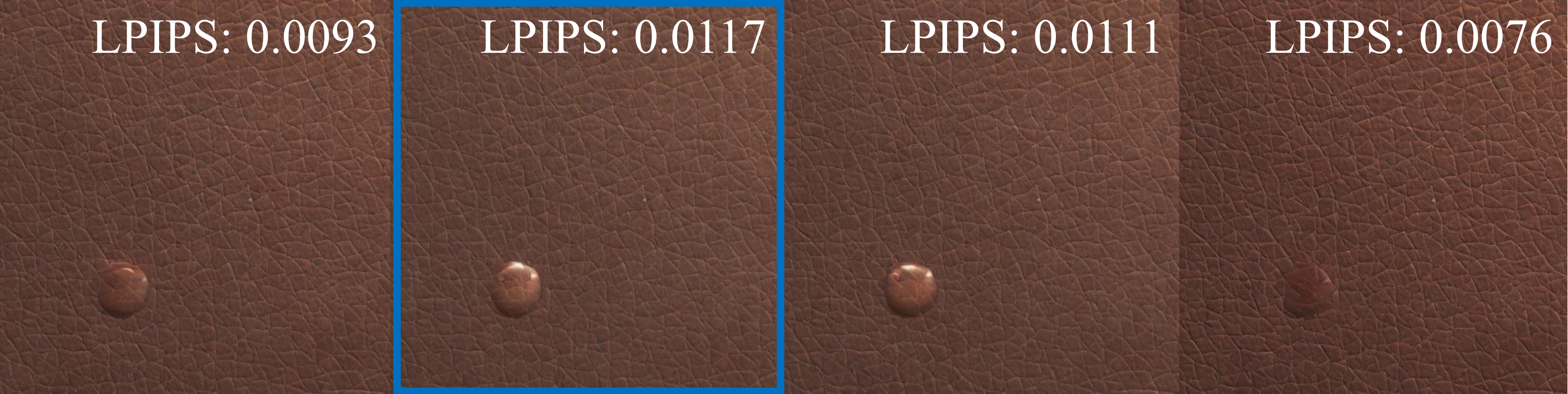} 
  \caption{\textbf{Low-Fidelity Selection (LFS) for defect of leather's glue.} LFS automatically selects the defect image with the most pronounced expression (blue box) by identifying the sample with the lowest fidelity (highest LPIPS score) in the masked area.}
  \label{fig:seed}
\end{figure}

\subsection{Learning Defect}
\label{learn_defect}

We use a stable diffusion inpainting model~\cite{rombach2022high} to leverage its prior knowledge for seamlessly ``filling” masked areas with desired defects. To train the model to understand the concept of defects, we fine-tune it using a small set of reference defect images $I$ paired with defect masks $M$. This fine-tuning enables the model to associate the word token [$V^*$] with defects. Specifically, to efficiently learn various defects while avoiding overfitting~\cite{ham2024personalized}, we fine-tune only the text encoder $\tau_\theta$ and the attention layers by using LoRA~\cite{hu2021lora}.

More precisely, we aim to achieve three goals to effectively learn local defects:
(1) recognizing defects that are not the main object of the image but rather local features dependent on it,
(2) understanding the semantic relationship between the defect and the main object to ensure natural blending, and
(3) ensuring the word token [$V^*$] corresponds to the defect region of the object.
To achieve these goals, we propose three loss terms: defect, object, and attention loss, as illustrated in the overall training scheme shown in~\cref{fig:method overview}

\paragraph{Defect Loss.}
\label{defect_loss}

The key loss term, defect loss $\mathcal{L}_{def}$, directly learns the detailed features of the defect concept. By guiding the model to focus exclusively on the intrinsic features of the defect, it enables inpainting of even unusual features that would not typically appear in the object.

First, we sample a timestep $t \sim p(t)$ from the model's timestep distribution and obtain the noise latent $x_t$ using~\cref{eq:x_t}. Next, we prepare the defect mask $M$ for $I$ and generate a background image $B_{def} = (1-M)\odot I$ where the defect area is masked out. The latent $b_{def} = \mathcal{E}(B_{def})$ is then concatenated with $x_t$ and $M$ to form $x_{t}^{def}$:
\begin{equation}
    x_{t}^{def}=\mathrm{concat}\left(x_t,b_{def},M\right)
    \label{eq:x_defect},
\end{equation}
which serves as input to the model.

To ensure the prompt focuses exclusively on the defect, we define it as $\mathcal{P}_{def} = \textit{``A photo of [$V^*$]"}$. The text encoder $\tau_\theta$ encodes this prompt to generate the text condition embedding $c^{def} = \tau_\theta(\mathcal{P}_{def})$. Using these inputs, we optimize the $\mathcal{L}_2$ loss with respect to noise $\epsilon$ to reconstruct $x_0$, but we compute the loss only over the masked region $M$ to avoid reconstructing the background:
\begin{equation}
    \mathcal{L}_{def} = \mathbb{E}_{x_{t}^{def}, t, \epsilon}\left[\left\|M \odot (\epsilon - \epsilon_\theta(x_{t}^{def}, t, c^{def}))\right\|_2^2\right].
    \label{eq:loss_defect}
\end{equation}

\paragraph{Object Loss.}

The object loss $\mathcal{L}_{obj}$ learns both the defect and its relationship to the object in which it appears. This ensures the defect blends naturally within the object.

The $\mathcal{L}_{obj}$ term shares the same sampled values for $\epsilon$, $t$, and $x_t$ as the defect loss. To capture the full semantic context of the object, we create a mask with 30 random boxes, $M_{rand}$, and train the model to fill in the occluded information across the entire image.
Similar to the defect loss, we obtain the conditioning background $B_{rand} = (1-M_{rand})\odot I$ and its latent $b_{rand}=\mathcal{E}(B_{rand})$. This $b_{rand}$ is then concatenated with $x_t$ and $M_{rand}$ to form $x_{t}^{obj}$:
\begin{equation}
  x_{t}^{obj}=\mathrm{concat}\left(x_t,b_{rand},M_{rand}\right)
  \label{eq:x_object}.
\end{equation}

To express the object's possession of the defect, we set the prompt as $\mathcal{P}_{obj} = \textit{``A [Object] with [$V^*$]"}$ and obtain the text embedding $c^{obj} = \tau_\theta(\mathcal{P}_{obj})$. Although it is essential to learn the semantic context of the defect within the object, capturing the fine details of the defect itself is also crucial for authentic inpainting. To address this, we apply a weight of 1 to the defect mask areas and a weight of $\alpha$, less than 1, to the background areas, producing an adjusted mask $M'$:
\begin{equation}
\begin{split}
    \mathcal{L}_{obj} &= \mathbb{E}_{x_{t}^{obj}, t, \epsilon}\left[\left\|M' \odot (\epsilon - \epsilon_\theta(x_{t}^{obj}, t, c^{obj}))\right\|_2^2\right], \\
    M' &= M + \alpha \cdot (1 - M).
\end{split}
\label{eq:loss_object}
\end{equation}

\paragraph{Attention Loss.}

We also utilize cross-attention maps from the forward pass for $\mathcal{L}_{obj}$. The maps for a specific token represent the layout of the corresponding object, allowing the model to focus more precisely on that region. This helps the model better attend to the defect’s features, resulting in higher-fidelity defect generation. Since the encoder in the UNet~\cite{ronneberger2015u} does not effectively represent the layout of the corresponding token object~\cite{cao2023masactrl}, we use only decoder-layer maps. To handle varying spatial sizes across decoder layers, we resize them to match the latent size, then average those of the [$V^*$] token to obtain $A_{t}^{[V^*]}$. Finally, we compute the $\mathcal{L}_2$ loss with the defect mask $M$, increasing values in the defect region while reducing them in the background:
\begin{equation}
    \mathcal{L}_{attn} = \mathbb{E}\left[\left\|A_{t}^{[V^*]} - M\right\|_2^2\right].
    \label{eq:loss_attn}
\end{equation}

\paragraph{DefectFill Loss.}
Finally, we fine-tune the model using a linear combination of these three loss terms:
\begin{equation}
    \mathcal{L}_{ours} =\lambda_{def}\cdot \mathcal{L}_{def} + \lambda_{obj}\cdot \mathcal{L}_{obj} + \lambda_{attn}\cdot \mathcal{L}_{attn}.
    \label{eq:loss_defectfill}
\end{equation}
The weights for each term are set to 0.5, 0.2, and 0.05, based on experiments that account for the scale of each loss. 

\begin{figure}[t!]
  \centering
  \includegraphics[width=\columnwidth]{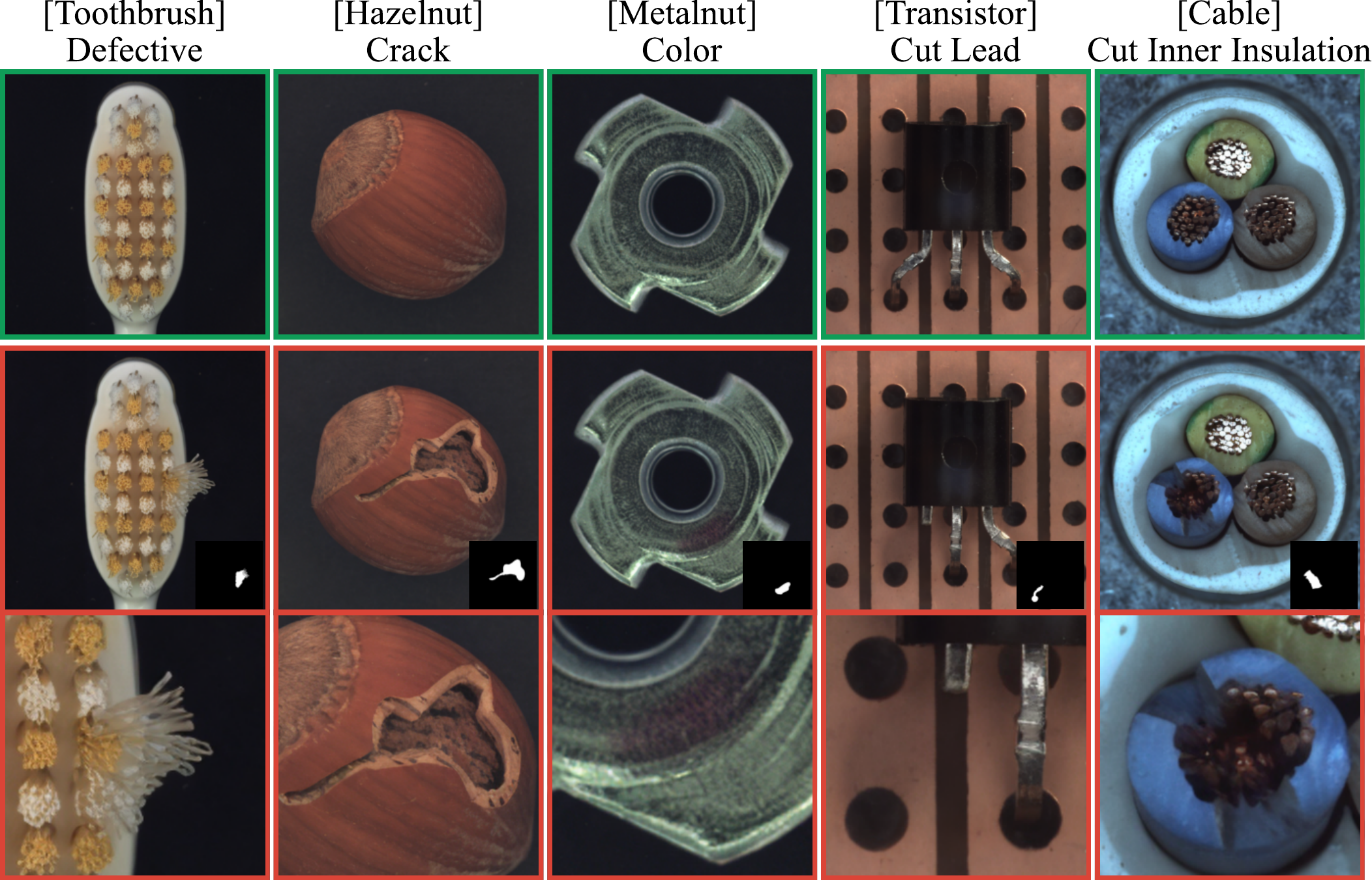} 
  \caption{\textbf{Generated Defects by DefectFill.} The first row displays the normal images (green boxes), while the second row shows the generated defect images along with their masks, and the third row provides zoomed-in views of the defects (red boxes). The zoomed images highlight the realistic and detailed rendering of the defects.}
  \label{fig:samples}
\end{figure}

\subsection{Generating Defect}
\label{generate_defect}

\paragraph{Sampling.}

After fine-tuning the inpainting diffusion model with our DefectFill loss to learn the defect concept, we utilize a widely adopted diffusion-based inpainting pipeline~\cite{avrahami2023blended,songscore} to generate diverse defect samples. Specifically, as input, we provide a defect-free image $I$ along with a mask $M$ indicating the exact area intended for defect placement. At each inference step $t$, we replace the latent representation's background area outside the mask with the latent of the defect-free image diffused with \cref{eq:x_t}. This approach ensures that the model modifies only the masked region while preserving the background that should remain unchanged. This approach maintains the structure of the original image, allowing for seamless integration of defects without affecting the overall image quality.

\paragraph{Low-Fidelity Selection.}

Finally, we propose an additional method for selecting samples where the defect is more accurately filled. Since the diffusion model generates diverse samples depending on the initial latent inputs, and due to the nature of the inpainting diffusion model, the masked area is occasionally overly reconstructed, resulting in lower-quality defect. To mitigate this issue, we select the least reconstructed image from the eight samples generated using the same normal image $I$ and defect mask $M$. This selection is based on a reconstruction metric (\eg PSNR, SSIM~\cite{ssim}, LPIPS~\cite{zhang2018unreasonable}) measured only within the masked region (as shown in~\cref{fig:seed}). This simple yet effective process filters out unclear cases and improves defect generation quality. In particular, for downstream tasks using generated defect images, this approach allows us to automatically select high-quality defects samples without manual effort. In our case, we employ LPIPS as the reconstruction metric.

\subsection{Applying to Visual Inspection}
\label{applying_downstream}

The generated high-quality defect images are used to train a visual inspection model. First, we learn the concept for each defect category (Sec.\ref{learn_defect}) and generate defective images (Sec.\ref{generate_defect}) for each category. After that, for classification, we train standard classification models (\eg ResNet~\cite{he2016deep}) using the generated images labeled by defect category. For localization, we train segmentation models (\eg UNet~\cite{ronneberger2015u}) with normal and synthesized defect images along with their corresponding masks, optimizing with focal loss~\cite{lin2017focal}.
\begin{figure*}[t!]
  \centering
  \includegraphics[width=0.95\textwidth]{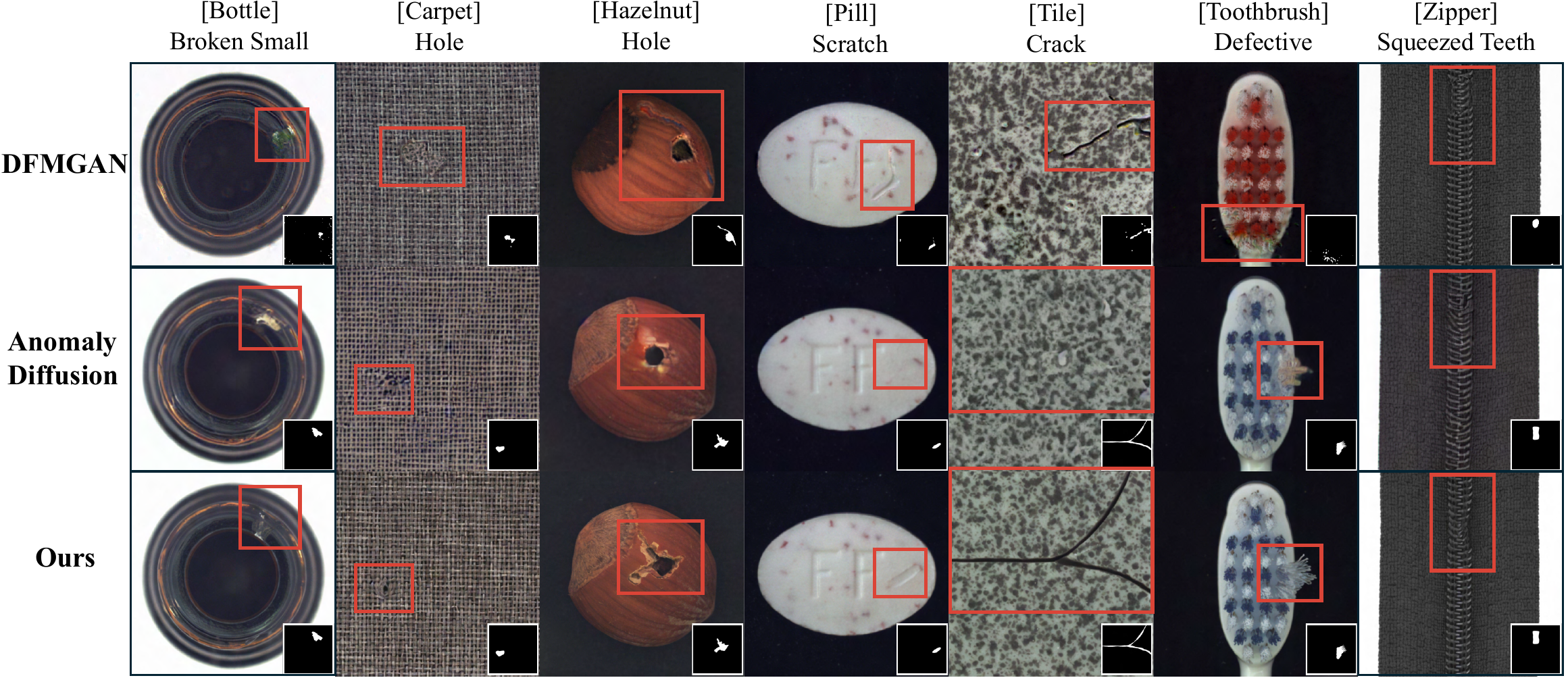} 
  \caption{\textbf{Defect Generation Comparisons.} This figure compares the quality of defect images generated by our method (bottom row) with baseline approaches. Our method produces the most realistic results, with defects that blend seamlessly into the objects.}
  \label{fig:Comparisons}
\end{figure*}

\begin{table*}[!htb] 
    \centering
    
    \begin{tabular}{c| c c | c c | c c | c c}
        \toprule
        \multirow{2}{*}{Objects} &\multicolumn{2}{c}{DFMGAN*} &\multicolumn{2}{c}{DFMGAN†} &\multicolumn{2}{c}{AnoDiff‡} &\multicolumn{2}{c}{Ours} \\\cmidrule{2-9}
        &KID↓ &IC-LPIPS↑ &KID↓ &IC-LPIPS↑ &KID↓ &IC-LPIPS↑ &KID↓ &IC-LPIPS↑ \\\midrule
        bottle &70.90 &0.12 &76.75 &0.15 &131.52 &\textbf{0.17} &\textbf{30.99} &0.12 \\
        capsule &40.63 &0.10 &182.83 &0.17 &54.97 &\textbf{0.18} &\textbf{5.60} &\textbf{0.18} \\
        carpet &\textbf{25.14} &0.13 &46.02 &\textbf{0.25} &149.83 &0.22 &50.37 &0.22 \\
        hazelnut &21.16 &0.24 &30.94 &\textbf{0.33} &50.61 &0.31 &\textbf{1.13} &0.31 \\
        leather &75.85 &0.17 &501.61 &\textbf{0.51} &244.75 &0.40 &\textbf{74.66} &0.30 \\
        pill &123.52 &0.16 &70.15 &\textbf{0.23} &77.69 &\textbf{0.23} &\textbf{8.76} &\textbf{0.23} \\
        tile &85.28 &0.22 &254.82 &0.25 &314.30 &\textbf{0.48} &\textbf{45.14} &0.44 \\
        toothbrush &46.49 &0.18 &61.43 &\textbf{0.20} &73.31 &0.18 &\textbf{3.19} &0.15 \\
        wood &68.13 &0.34 &406.61 &\textbf{0.35} &83.94 &\textbf{0.35} &\textbf{4.72} &\textbf{0.35} \\
        zipper &78.08 &\textbf{0.27} &35.74 &0.26 &126.65 &0.24 &\textbf{34.91} &0.20 \\
        \bottomrule
    \end{tabular}

    \caption{\textbf{Generation Comparison.} This table presents the average KID and IC-LPIPS scores, computed from 1,000 generated images per defect category and averaged across all categories for each object. Our method achieves the best KID scores for all objects except carpet and the highest IC-LPIPS scores for capsule, pill, and wood. DFMGAN*: scores taken directly from the paper. DFMGAN†: scores reproduced by us. AnoDiff‡: scores measured from the generated dataset (poor samples filtered) on their official page.}
    \label{tab:generation}
\end{table*}

\section{Experiments}

\paragraph{Dataset.}  

We evaluate DefectFill on the MVTec AD Dataset~\cite{bergmann2019mvtec}, which consists of 15 industrial objects with multiple defect categories. Each category contains hundreds of normal images and approximately 20 defect images with masks. Instead of traditional anomaly detection, we generate defect images by training on one-third of the defect image-mask pairs and applying the model to the remaining two-thirds of masks with normal images. For reliable quantitative results, we evaluate on 10 objects, while all objects are used for qualitative analysis.

\paragraph{Implementation Details.}  

Our approach leverages the Stable-Diffusion-2-inpainting model~\cite{rombach2022high}, fine-tuning the text encoder and UNet's attention layers with LoRA (rank 8)~\cite{hu2021lora}. We use a learning rate of 2e-4 for the UNet and 4e-5 for the text encoder. Inference is conducted with a DDIM~\cite{song2020denoising} scheduler with 50 denoising steps. Additional details are provided in the supplementary materials.

\paragraph{Metric.}  

We evaluate defect generation quality using Kernel Inception Distance (KID)~\cite{binkowski2018demystifying} for quality and IC-LPIPS~\cite{ojha2021few} for diversity, excluding FID and IS due to their limitations on smaller or unreferenced datasets. For defect inspection, we measure classification accuracy, Area Under the ROC Curve (AUROC), Average Precision (AP), $F_1$-max, and Per Region Overlap (PRO).

\paragraph{Baselines.}  

We compare our method against two state-of-the-art defect generation methods: DFMGAN~\cite{duan2023few}, a two-stage GAN-based approach, and AnomalyDiffusion~\cite{hu2024anomalydiffusion}, a text-to-image diffusion model that disentangles the appearance and spatial attributes of defects.

\setlength{\tabcolsep}{3pt} 

\begin{table*}[!htb]
    \centering

    \begin{adjustbox}{max width = 1.0\linewidth}
    \begin{tabular}{c|cccc|cccc|cccc|cccc}
        \toprule
        \multirow{2}{*}{Objects} &\multicolumn{4}{c}{DFMGAN†} &\multicolumn{4}{c}{AnoDiff*} &\multicolumn{4}{c}{AnoDiff‡} &\multicolumn{4}{c}{Ours} \\\cmidrule{2-17}
        &AUROC &AP &$F_1$-max &PRO &AUROC &AP &$F_1$-max &PRO &AUROC &AP &$F_1$-max &PRO &AUROC &AP &$F_1$-max &PRO \\\midrule
        bottle &0.96 &0.80 &0.74 &0.84 &0.99 &0.94 &0.87 &0.94 &0.99 &0.91 &0.83 &0.94 &\textbf{1.00} &\textbf{0.96} &\textbf{0.90} &\textbf{0.97} \\
        capsule &0.74 &0.04 &0.09 &0.34 &0.99 &0.57 &0.60 &0.95 &0.98 &0.41 &0.44 &0.84 &\textbf{1.00} &\textbf{0.75} &\textbf{0.69} &\textbf{0.96} \\
        carpet &0.95 &0.62 &0.60 &0.84 &\textbf{0.99} &0.81 &0.75 &0.92 &0.97 &0.74 &0.68 &0.82 &\textbf{0.99} &\textbf{0.92} &\textbf{0.86} &\textbf{0.96} \\
        hazelnut &\textbf{1.00} &0.94 &0.87 &0.96 &\textbf{1.00} &0.97 &0.91 &0.97 &\textbf{1.00} &0.96 &0.90 &0.97 &\textbf{1.00} &\textbf{0.99} &\textbf{0.94} &\textbf{0.99} \\
        leather &0.96 &0.56 &0.56 &0.84 &\textbf{1.00} &0.80 &0.71 &\textbf{0.98} &\textbf{1.00} &0.80 &0.72 &\textbf{0.98} &\textbf{1.00} &\textbf{0.91} &\textbf{0.83} &\textbf{0.98} \\
        pill &0.99 &0.89 &0.86 &0.91 &\textbf{1.00} &0.97 &0.91 &0.97 &\textbf{1.00} &0.97 &0.91 &0.95 &\textbf{1.00} &\textbf{0.98} &\textbf{0.93} &\textbf{0.98} \\
        tile &0.99 &0.94 &0.87 &0.96 &0.99 &0.94 &0.86 &0.96 &0.99 &0.95 &0.87 &0.97 &\textbf{1.00} &\textbf{0.97} &\textbf{0.90} &\textbf{0.98} \\
        toothbrush &0.98 &0.60 &0.61 &0.89 &\textbf{0.99} &0.77 &0.73 &0.91 &\textbf{0.99} &0.71 &0.68 &0.91 &\textbf{0.99} &\textbf{0.89} &\textbf{0.82} &\textbf{0.94} \\
        wood &0.72 &0.37 &0.41 &0.73 &0.99 &0.85 &0.75 &0.94 &0.97 &0.79 &0.72 &0.89 &\textbf{1.00} &\textbf{0.93} &\textbf{0.86} &\textbf{0.98} \\
        zipper &0.99 &0.82 &0.76 &0.95 &0.99 &0.86 &0.79 &0.96 &\textbf{1.00} &0.87 &0.80 &\textbf{0.97} &\textbf{1.00} &\textbf{0.90} &\textbf{0.84} &\textbf{0.97} \\
        \bottomrule
    \end{tabular}
    \end{adjustbox}

    \caption{\textbf{Localization Comparison.} The table presents AUROC, AP, $F_1$-max, and PRO scores for localization evaluation using a UNet trained on generated defect images. Our method achieves the highest performance across all metrics and objects. AnoDiff*: scores reported in the paper. The others: described in~\cref{tab:generation}.
    }
    \label{tab:localization}
\end{table*}

\subsection{Defect Generation Evaluation}
\label{generation_results} 

\paragraph{Qualitative Results.}  
\label{qualitative_results} 

The generated results are shown in~\cref{fig:samples}. The first row displays the normal images, the second row shows the generated defect images using the mask in the lower right, and the third row provides a zoomed-in view of the generated defects. Despite using custom-drawn masks which are unseen during training, the model generates authentic and well-aligned defects. Notably, for the hazelnut, the model produces a realistic defect that aligns with the object’s semantics, even with an unrealistic mask shape for the crack, demonstrating its strong generalization ability. Additionally, the detailed texture within the hazelnut is observable and highlights the realism of the defects.

\cref{fig:Comparisons} presents a qualitative comparison with the baselines. For AnomalyDiffusion, we use the same normal image and mask, while DFMGAN cannot use the same base image as it generates both normal and defect images directly. In the hazelnut case, both baselines struggle with the texture around the hole, whereas our method produces realistic defects that blend seamlessly with the object texture, handling irregular mask shapes and demonstrating DefectFill's robustness. In cases such as carpet and tile, where defects are small or thin, the baselines either fail to capture them accurately or omit them entirely, while our model generates well-defined defects. For the toothbrush, DFMGAN blurs the masked area, and AnomalyDiffusion generates defects with colors misaligned with the object context. In contrast, our model produces a realistic blueish defect that reflects the object's context (similar to how the toothbrush in~\cref{fig:samples} appears yellowish). This demonstrates our model's ability to integrate object semantics into defect generation.

\setlength{\tabcolsep}{3.5pt} 

\begin{table}[!tb]
    \centering
    
    \begin{adjustbox}{max width = 1.0\linewidth} 
    \begin{tabular}{c|ccccc}
        \toprule
        Objects &DFMGAN* &DFMGAN† &AnoDiff* &AnoDiff‡ &Ours \\\midrule
        bottle &56.59 &63.41 &90.70 &95.35 &\textbf{97.56} \\
        capsule &37.27 &25.00 &66.67 &45.33 &\textbf{87.50} \\
        carpet &47.31 &42.11 &58.06 &64.52 &\textbf{87.72} \\
        hazelnut &81.94 &86.96 &85.42 &89.58 &\textbf{100.00} \\
        leather &49.73 &32.20 &61.90 &65.08 &\textbf{93.22} \\
        pill &29.52 &44.44 &59.38 &64.58 &\textbf{97.53} \\
        tile &74.85 &81.82 &84.21 &96.49 &\textbf{100.00} \\
        wood &49.02 &45.16 &71.43 &78.57 &\textbf{100.00} \\
        zipper &27.64 &45.45 &69.51 &85.37 &\textbf{90.91} \\
        \bottomrule
    \end{tabular}
    \end{adjustbox}
    
    \caption{\textbf{Classification Comparison.} The table shows classification accuracy (\%) when a ResNet-34 is trained on generated defect images for defect category prediction. Our method achieves the highest performance across all objects. AnoDiff*: scores reported in the paper. The others: described in~\cref{tab:generation}.}

    \label{tab:classification}
    
\end{table}

\paragraph{Quantitative Results.}  
\label{Quantitative Results}

Tab.~\ref{tab:generation} compares the KID and IC-LPIPS scores of our method with baseline approaches across various objects.
For evaluation, we generate 1,000 images for each defect category within each object, ensuring that all metrics, including KID, are calculated using only defect images excluded from the training set. This approach is necessary as KID often produces overly optimistic values when models overfit and are evaluated on training data.

Our method outperforms the baselines in KID scores across most objects. For IC-LPIPS, it also achieves the best scores on three objects (capsule, pill, wood). In the case of leather, DFMGAN† and AnoDiff‡ score significantly higher, but this is primarily due to their generation of diverse yet low-quality samples across various masks. The high KID values for these methods further confirm that the quality of their generated defect images is low.

\subsection{Visual Inspection Evaluation}
\label{inspection} 

To demonstrate that the realistic images generated by DefectFill can enhance performance in downstream visual inspection tasks, we apply it to two tasks: classification and localization. Following the experimental setup of AnomalyDiffusion~\cite{hu2024anomalydiffusion}, we use ResNet-34~\cite{he2016deep} for classification and UNet~\cite{ronneberger2015u} for localization. As outlined in the quantitative results (\cref{Quantitative Results}), we generate 1,000 defects per category and train the models on this data. Testing is conducted on the remaining two-thirds of the dataset.

\paragraph{Classification.}

As shown in Tab.\ref{tab:classification}, our method achieves higher classification accuracy across all objects compared to the baselines. Notably, there is a significant improvement for objects with small defect areas, which are typically challenging to generate meaningful defects for, such as capsule ($66.67\% \rightarrow 87.50\%$) and pill ($64.58\% \rightarrow 97.53\%$).

\paragraph{Localization.}

The UNet is trained to predict defect locations, and the predictions are evaluated using various metrics. As shown in \cref{tab:localization}, our model achieves the best performance across all metrics and objects. The capsule is a particularly challenging object for localization, yet our model significantly outperforms the baseline with a notable improvement in AP score ($0.57 \rightarrow 0.75$).

\begin{figure}[!tb]
  \centering
  \includegraphics[width=\columnwidth]{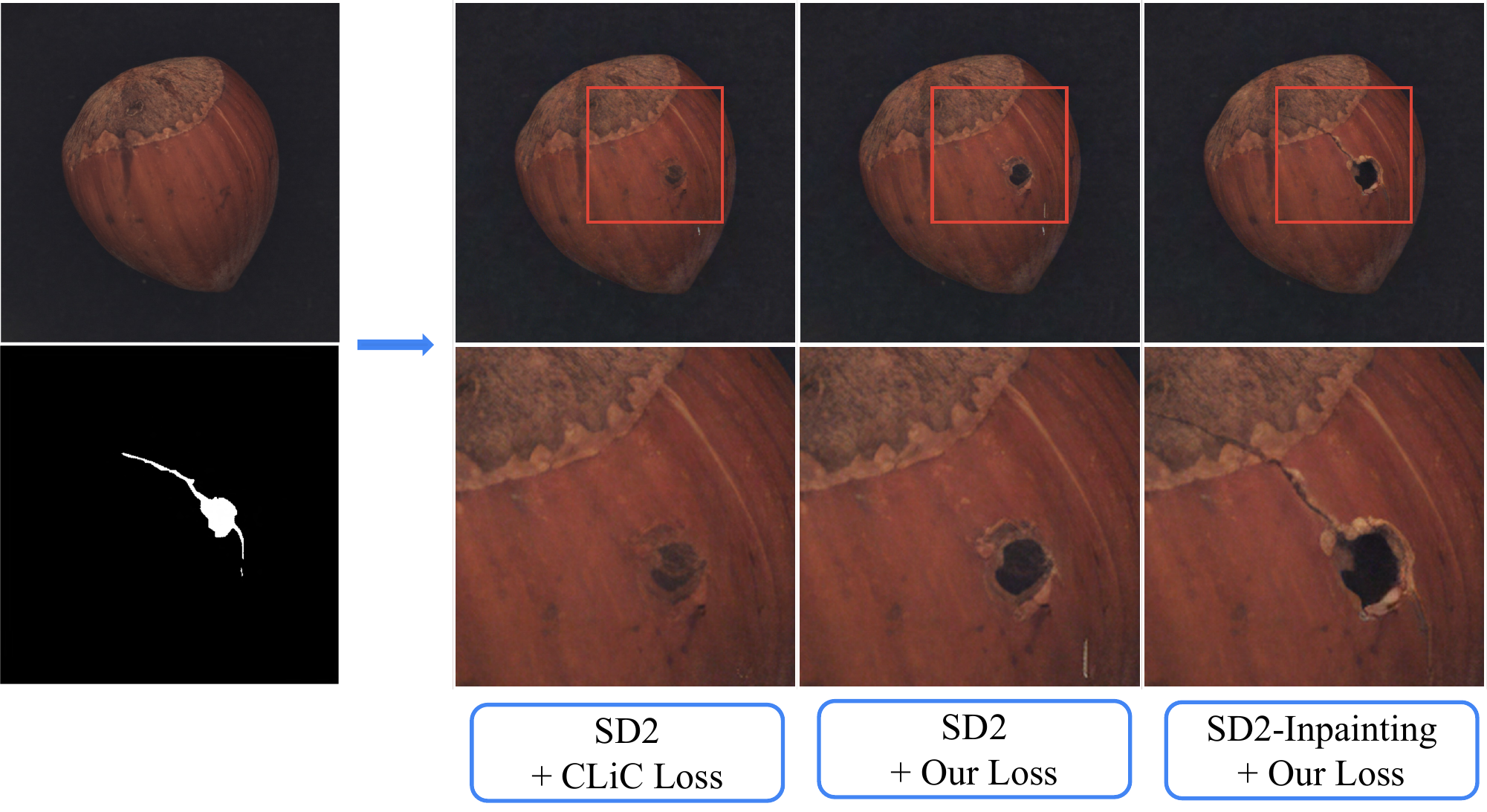} 
  \caption{\textbf{Inpainting Ablation.} Ablation study comparing three setups: applying CLiC~\cite{safaee2024clic} loss to vanilla Stable Diffusion (SD2+CLiC Loss), replacing CLiC with our loss (SD2+Our Loss), and our full approach (SD2-Inpainting+Our Loss). Using the inpainting model with our loss is necessary to produce realistic defects that align well with both the mask and the object.}
  \label{fig:inpainting ablation}
\end{figure}

\subsection{Ablation Studies}
\label{ablation} 

\paragraph{Inpainting Ablation.}   

We conduct an ablation study to evaluate the impact of leveraging the inpainting diffusion model and our defect-specific loss tailored for this model (\cref{fig:inpainting ablation}). As mentioned in \cref{sec:related_work}, CLiC~\cite{safaee2024clic} is a method that learns local concepts without using the inpainting diffusion model. However, the generated results tend to focus on reconstruction rather than creating actual holes (left image). This is because, unlike general local concepts, the defect concept we aim to learn is an unusual concept unknown to the model's prior. When applying our defect-specific loss (middle image), the model better learns the defect features, resulting in more accurately formed holes. However, the thin regions of the mask are still neglected, and the texture around the hole doesn’t blend well with the surrounding hazelnut texture. Finally, by leveraging the inpainting diffusion model’s strong prior for filling, we generate realistic defects that blend naturally with their surroundings (as shown by the light brown texture around the hole in the right image), and aligning with the thin mask regions.

\paragraph{Loss Ablation.}  

\begin{figure}[!tb]
  \centering
  \includegraphics[width=\columnwidth]{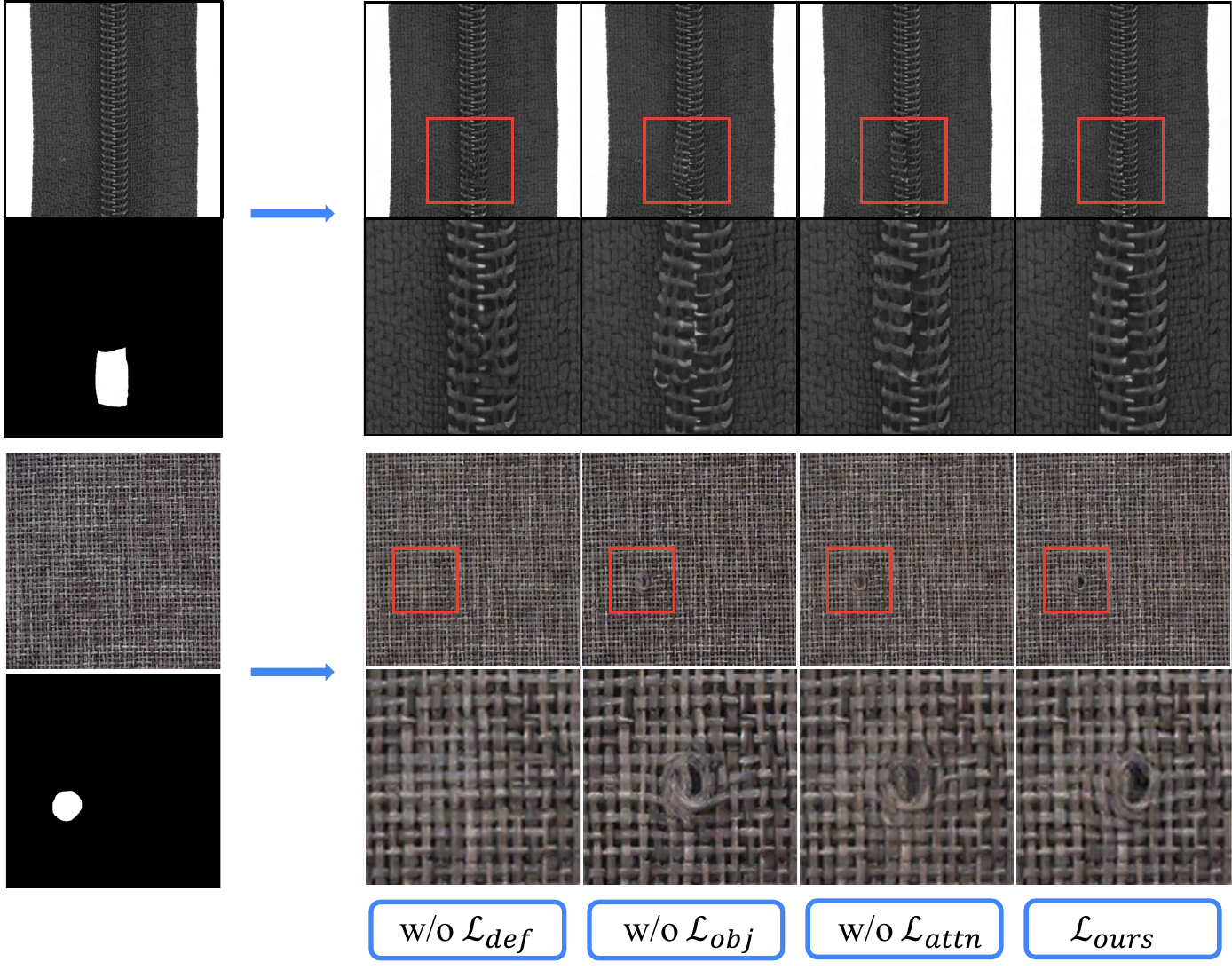} 
  \caption{\textbf{Loss Ablation.} This figure illustrates the impact of each loss term on generated defect quality. We show the results when each loss term is individually removed during fine-tuning, as well as the result when all terms are used together. Utilizing all loss terms results in realistic defects that align well with the context.}
  \label{fig:loss ablation}
\end{figure}

As described in \cref{learn_defect}, we structure our loss function with three terms to achieve three specific goals. To illustrate the contribution of each term, we perform an ablation study. \cref{fig:loss ablation} shows the defect generation results when each loss term is omitted during training. When the defect loss is excluded, the model tends to reconstruct rather than generate defects. This occurs because the inpainting diffusion model fails to learn the distinctive characteristics of defects and instead fills the masked area with just plausible context. Without the object loss, the model lacks semantic alignment with the object, leading to unnatural defect generation. For example, the middle section of a zipper may appear fused, or a hole may look like it's placed on a carpet rather than genuinely puncturing it. Lastly, when the attention loss is omitted, the model struggles to focus accurately on the defect mask area, resulting in lower defect fidelity (\eg an awkwardly split zipper or an incomplete hole). Finally, by combining all loss terms, we achieve realistic defects seamlessly filled onto objects.

\paragraph{Low Fidelity Selection.}  

Our simple yet effective Low-Fidelity Selection method enables high-quality defect sampling without human effort. As shown in \cref{fig:seed}, it intuitively selects qualitatively good samples. Additionally, as reported in \cref{tab:lfs_avg}, it improves both the quality (KID) and the diversity (IC-LPIPS) of generated defects.

\begin{table}[!tp]
\centering

    \begin{tabular}{cc|ccc}\toprule
    \multicolumn{2}{c}{Ours w/o LFS} &\multicolumn{2}{c}{Ours} \\\cmidrule{1-4}
    KID↓ &IC-LPIPS↑ &KID↓ &IC-LPIPS↑ \\\midrule
    28.485 &0.247 &\textbf{25.947} &\textbf{0.250} \\
    \bottomrule
    \end{tabular}
    \caption{\textbf{Generation Comparison with Low-Fidelity Selection.} The application of LFS shows improvements in quality (KID) and diversity (IC-LPIPS). The values represent averages calculated for each defect category, and then averaged across objects.}
    \label{tab:lfs_avg}
\end{table}
\section{Conclusions}
\label{sec:conclusion}
 
In this work, we present DefectFill, a novel approach that fine-tunes an inpainting diffusion model to generate realistic and high-fidelity defect images. Our method achieves state-of-the-art performance in both generation quality and visual inspection tasks on the MVTec AD dataset, demonstrating its effectiveness even when limited reference samples are available. These strengths make DefectFill particularly well-suited for widespread industrial applications, especially in scenarios where defect images are scarce.

\paragraph{Limitations.}

While our method excels at generating localized defects—a common real-world scenario—it is less effective for global structural defects that affect the entire object, such as misalignment. This limitation arises because our inpainting-based approach, which focuses on local masked regions. Addressing such global defects remains an area for future research, though our method already robustly handles the majority of practical defect cases, where localized defects are the primary concern.

\paragraph{Acknowledgement} This work was supported by the National Research Foundation of Korea (NRF) grant funded by the Korea government (MSIT) [No. 2022R1A3B1077720], Institute of Information \& communications Technology Planning \& Evaluation (IITP) grant funded by the Korea government(MSIT) [NO.RS-2021-II211343, Artificial Intelligence Graduate School Program (Seoul National University)], and the BK21 FOUR program of the Education and Research Program for Future ICT Pioneers, Seoul National University in 2024.

{
    \small
    \bibliographystyle{ieeenat_fullname}
    \bibliography{main}
}

\clearpage
\setcounter{section}{0}
\renewcommand\thesection{\Alph{section}}
\setcounter{table}{0}
\renewcommand{\thetable}{S\arabic{table}}
\setcounter{figure}{0}
\renewcommand{\thefigure}{S\arabic{figure}}
\section*{Appendix}

\section{Training Details}
\label{sec:training details}

We use a batch size of 4 for training. The learning rate is set to $2\times10^{-4}$ for the UNet~\cite{ronneberger2015u} and $4\times10^{-5}$ for the text encoder. Training is conducted over 2000 steps, with the first 100 steps dedicated to warmup, during which the learning rate linearly increases from 0 to its specified value. Throughout the training, images $I$ and masks $M$ are randomly resized together by a factor between $1.0$ and $1.125\times$ and then cropped back to their original size. Random masks are generated using 30 boxes with side lengths randomly chosen between 3\% and 25\% of the image size.
We fine-tune only the projection matrices of the text encoder and UNet using LoRA~\cite{hu2021lora} with a rank of 8. The dropout rate is set to 0.1, and the LoRA scaling factor is set to 16. For the [$V^*$] token, we use the word \textit{``sks"}.
For the DefectFill loss, we assign weights of 0.5, 0.2, and 0.05 to the defect loss, object loss, and attention loss, respectively. The adjusted mask $M'$ used in the object loss calculation has $\alpha$ value set to 0.3.

\section{Additional Qualitative Results}
\label{sec:additional qualitative results}

\subsection{MVTec AD Dataset}

\begin{figure*}
  \centering
    \includegraphics[width=\textwidth]{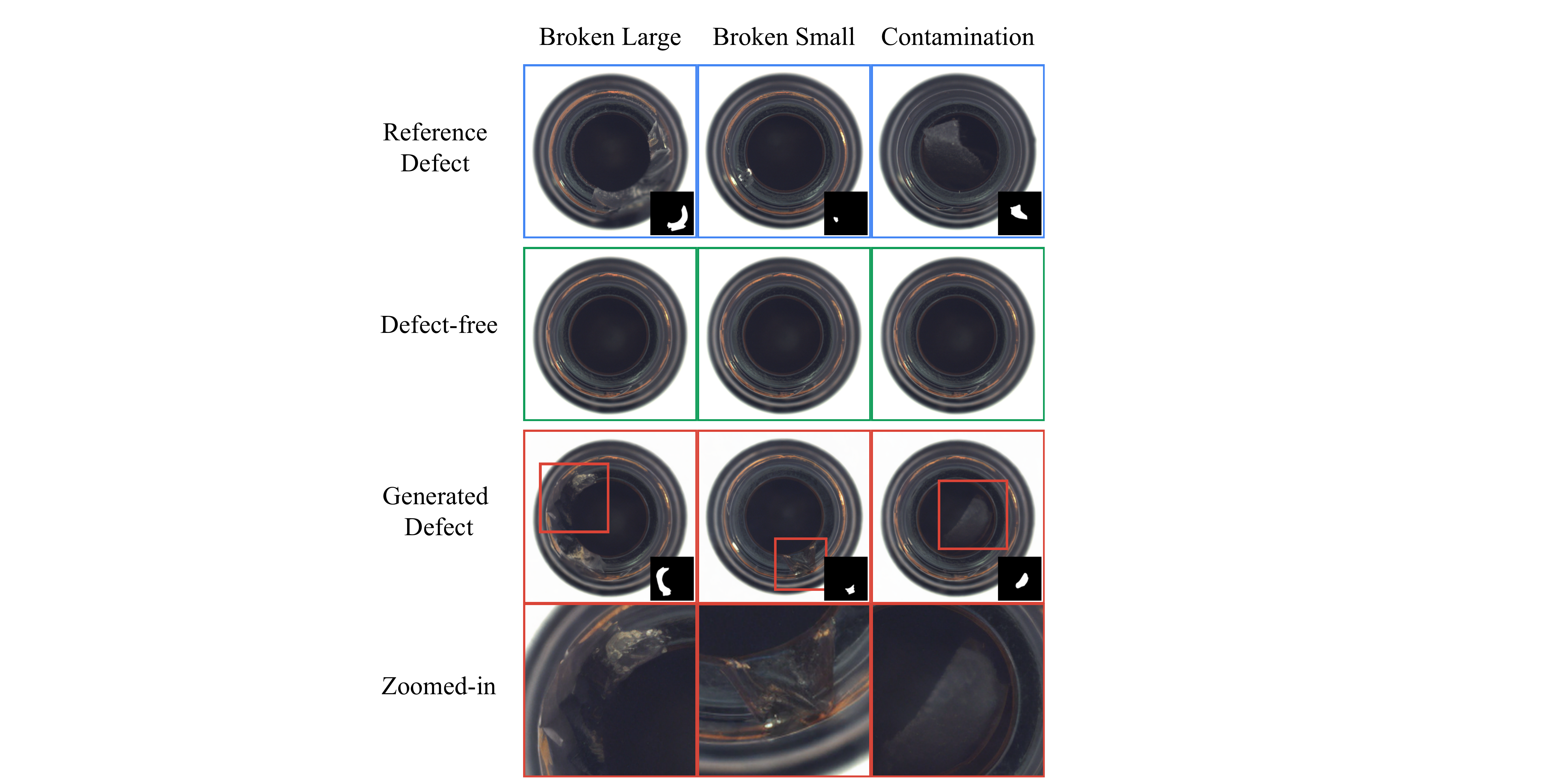}
    \caption{\textbf{Defect generation results on MVTec AD dataset (object: bottle).}}
    \label{fig:bottle}
  \hfill
\end{figure*}

\begin{figure*}
  \centering
    \includegraphics[width=\textwidth]{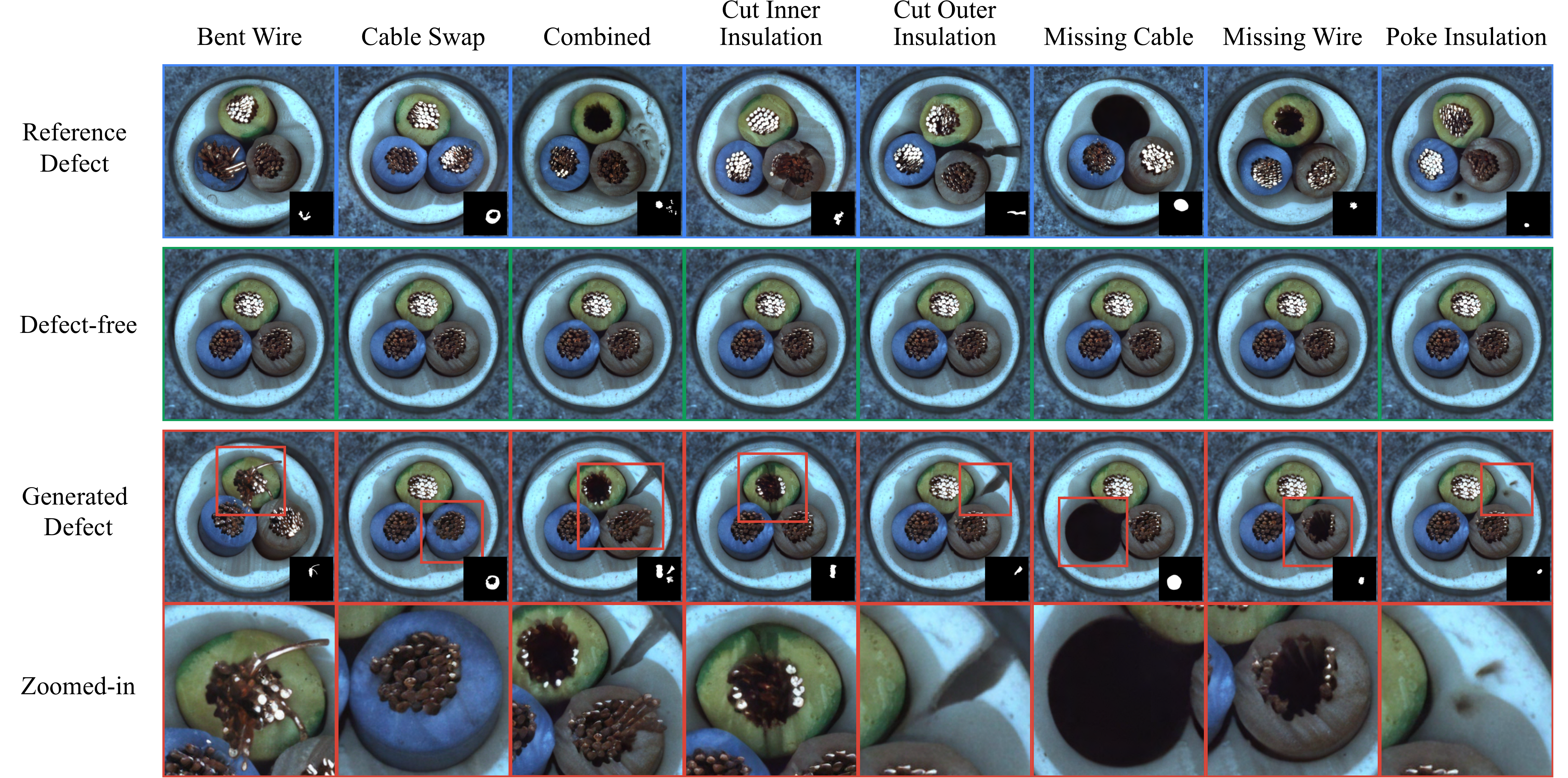}
    \caption{\textbf{Defect generation results on MVTec AD dataset (object: cable).}}
    \label{fig:cable}
  \hfill
\end{figure*}

\begin{figure*}
  \centering
    \includegraphics[width=\textwidth]{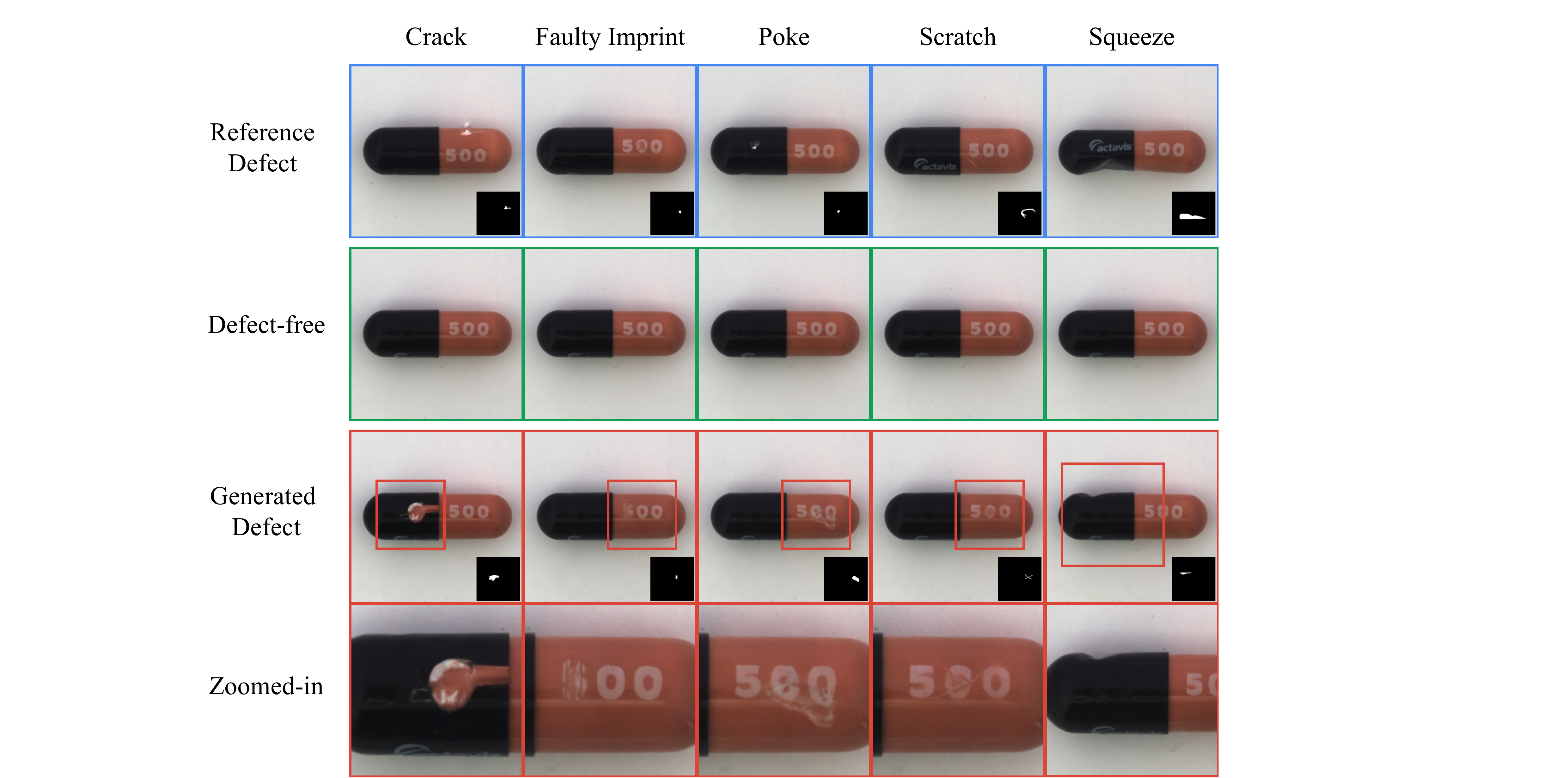}
    \caption{\textbf{Defect generation results on MVTec AD dataset (object: capsule).}}
    \label{fig:capsule}
  \hfill
\end{figure*}

\begin{figure*}
  \centering
    \includegraphics[width=\textwidth]{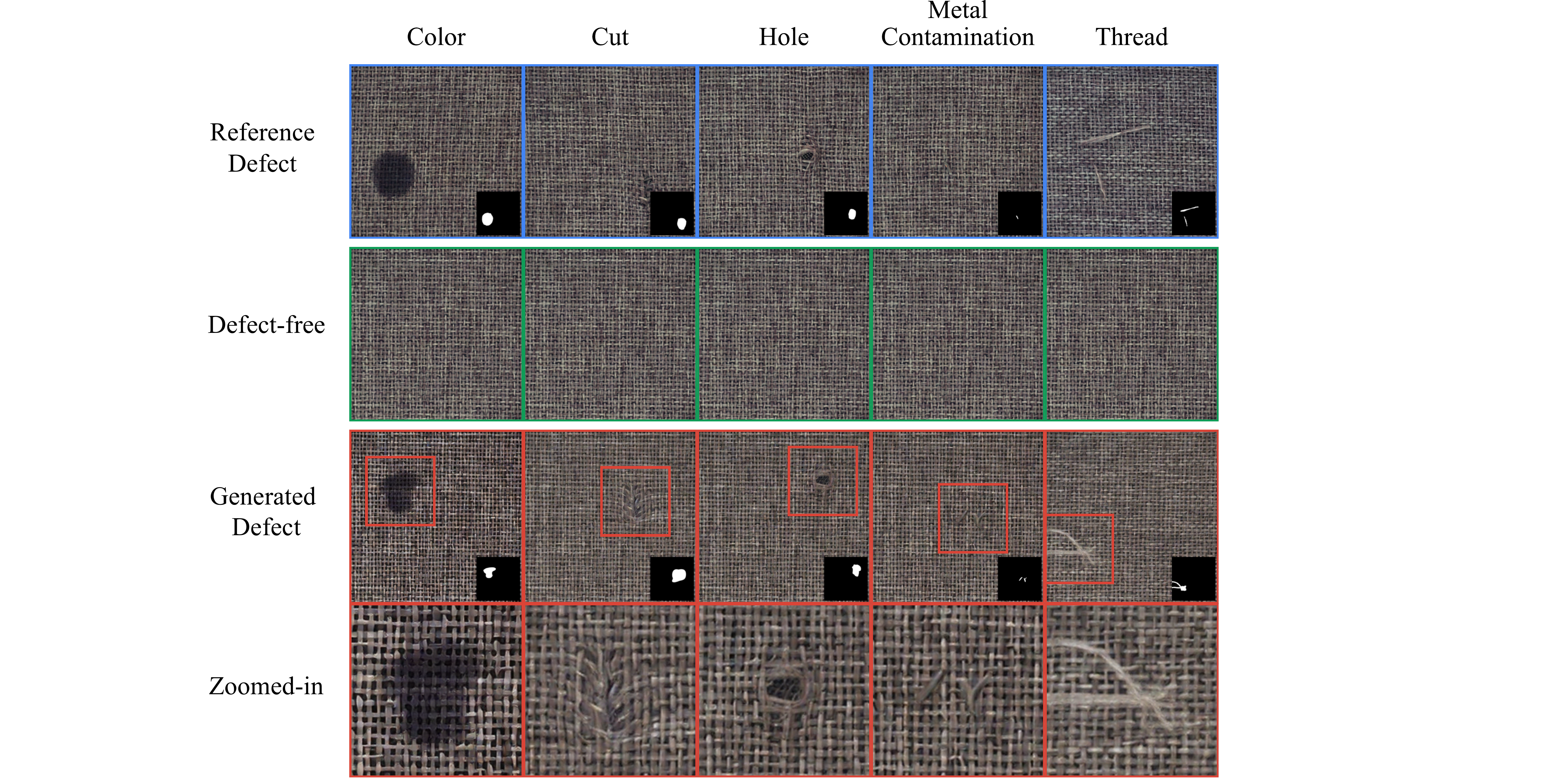}
    \caption{\textbf{Defect generation results on MVTec AD dataset (object: carpet).}}
    \label{fig:carpet}
  \hfill
\end{figure*}

\begin{figure*}
  \centering
    \includegraphics[width=\textwidth]{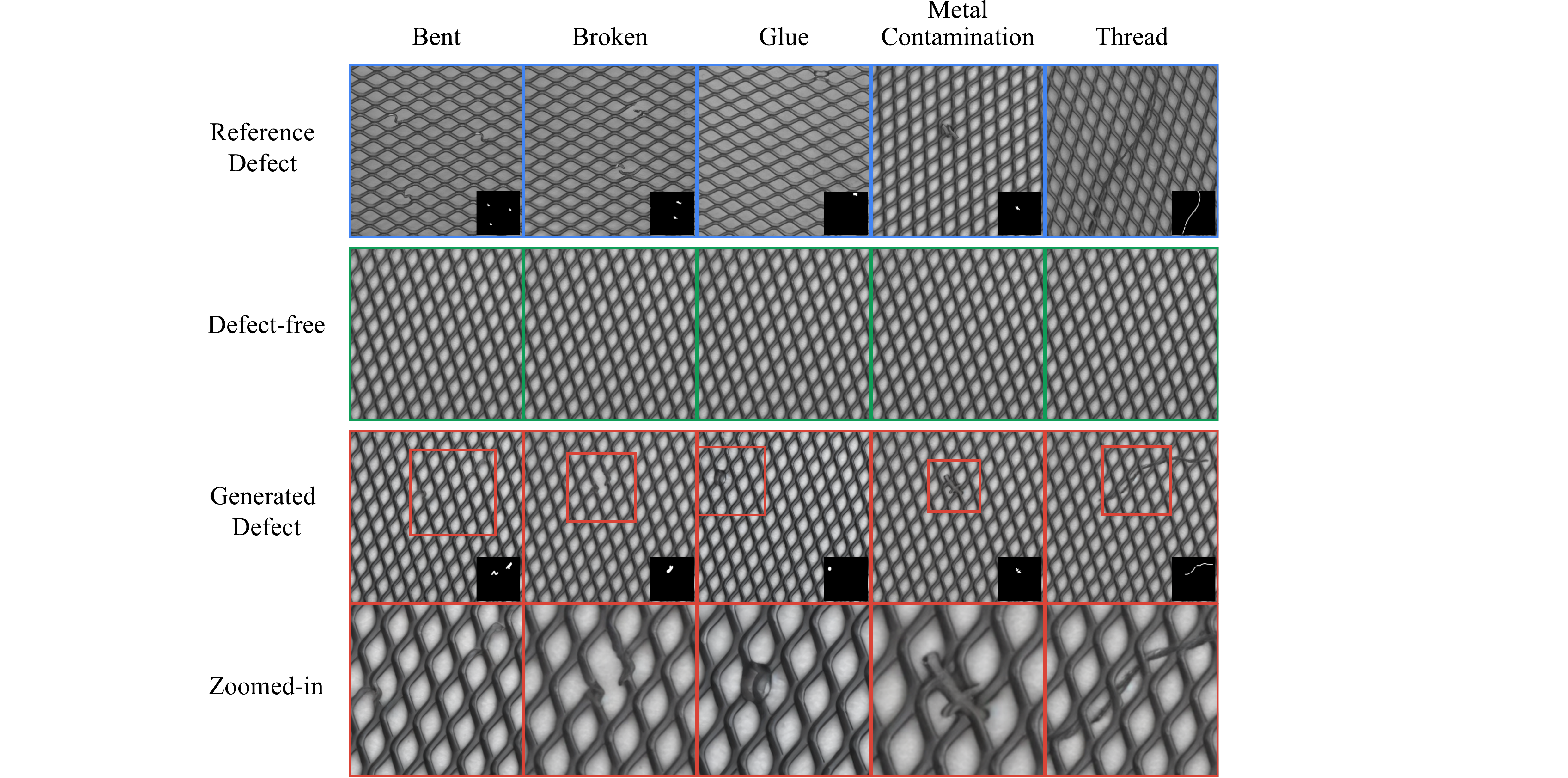}
    \caption{\textbf{Defect generation results on MVTec AD dataset (object: grid).}}
    \label{fig:grid}
  \hfill
\end{figure*}

\begin{figure*}
  \centering
    \includegraphics[width=\textwidth]{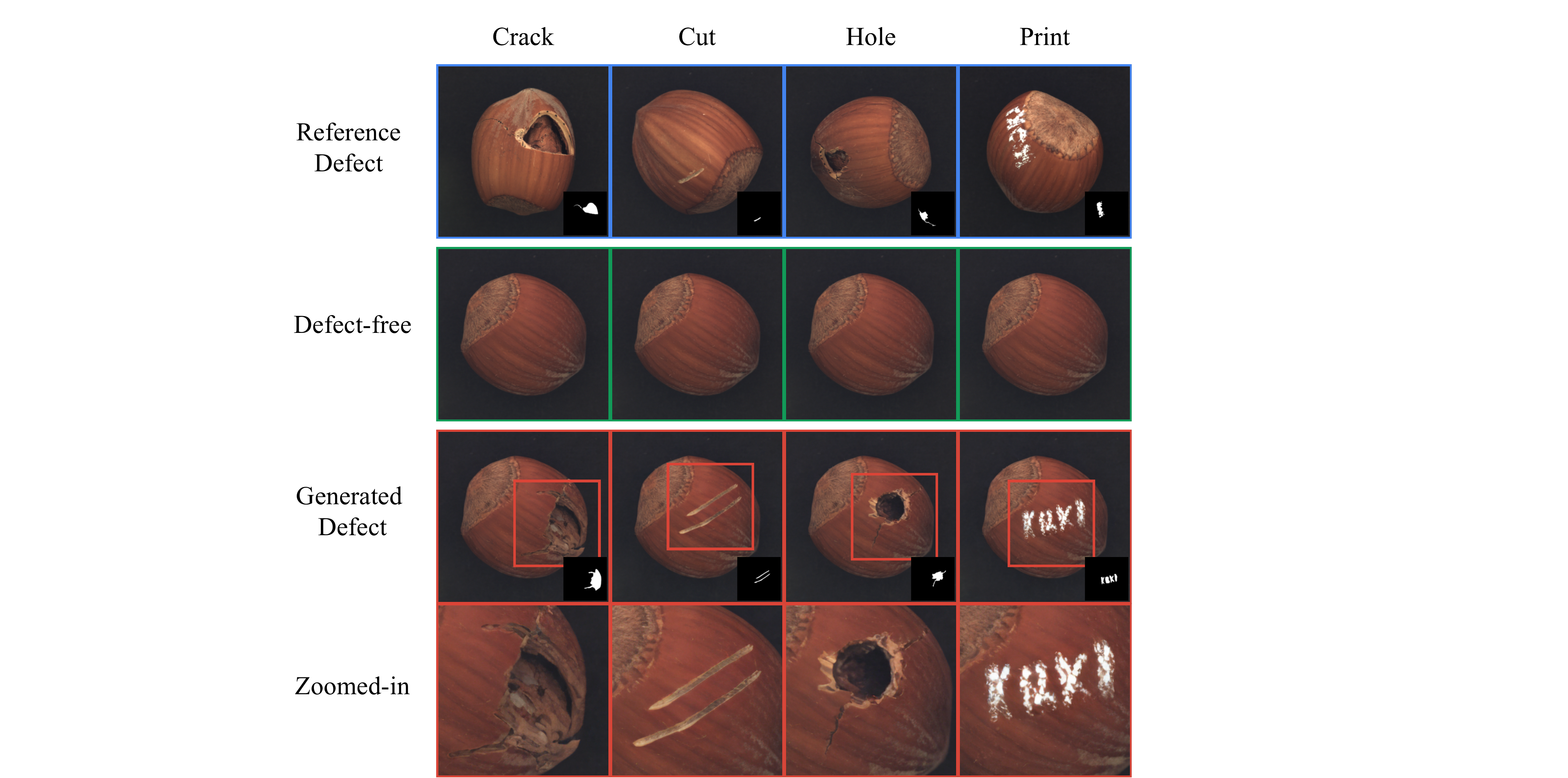}
    \caption{\textbf{Defect generation results on MVTec AD dataset (object: hazelnut).}}
    \label{fig:hazelnut}
  \hfill
\end{figure*}

\begin{figure*}
  \centering
    \includegraphics[width=\textwidth]{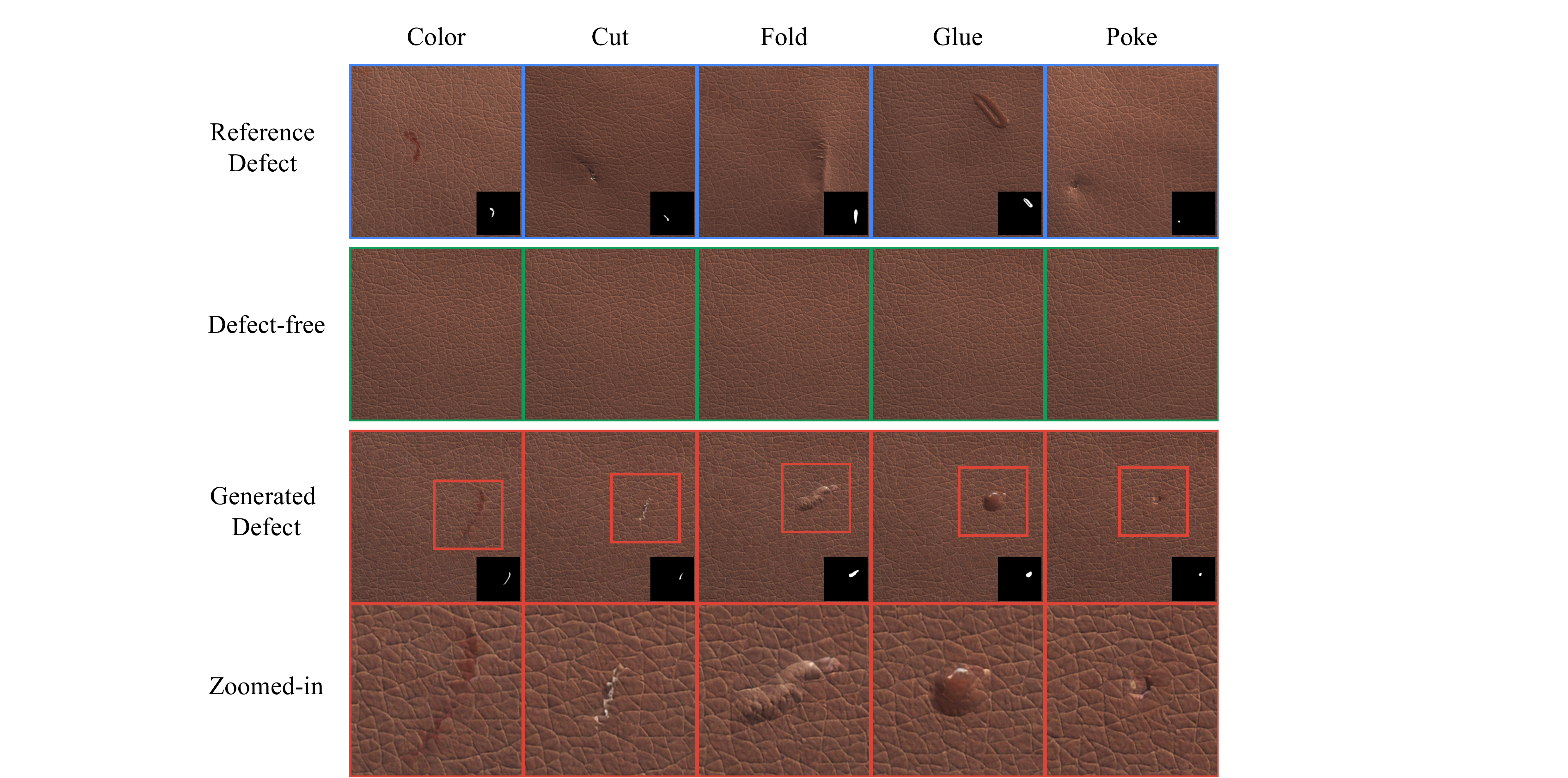}
    \caption{\textbf{Defect generation results on MVTec AD dataset (object: leather).}}
    \label{fig:leather}
  \hfill
\end{figure*}

\begin{figure*}
  \centering
    \includegraphics[width=\textwidth]{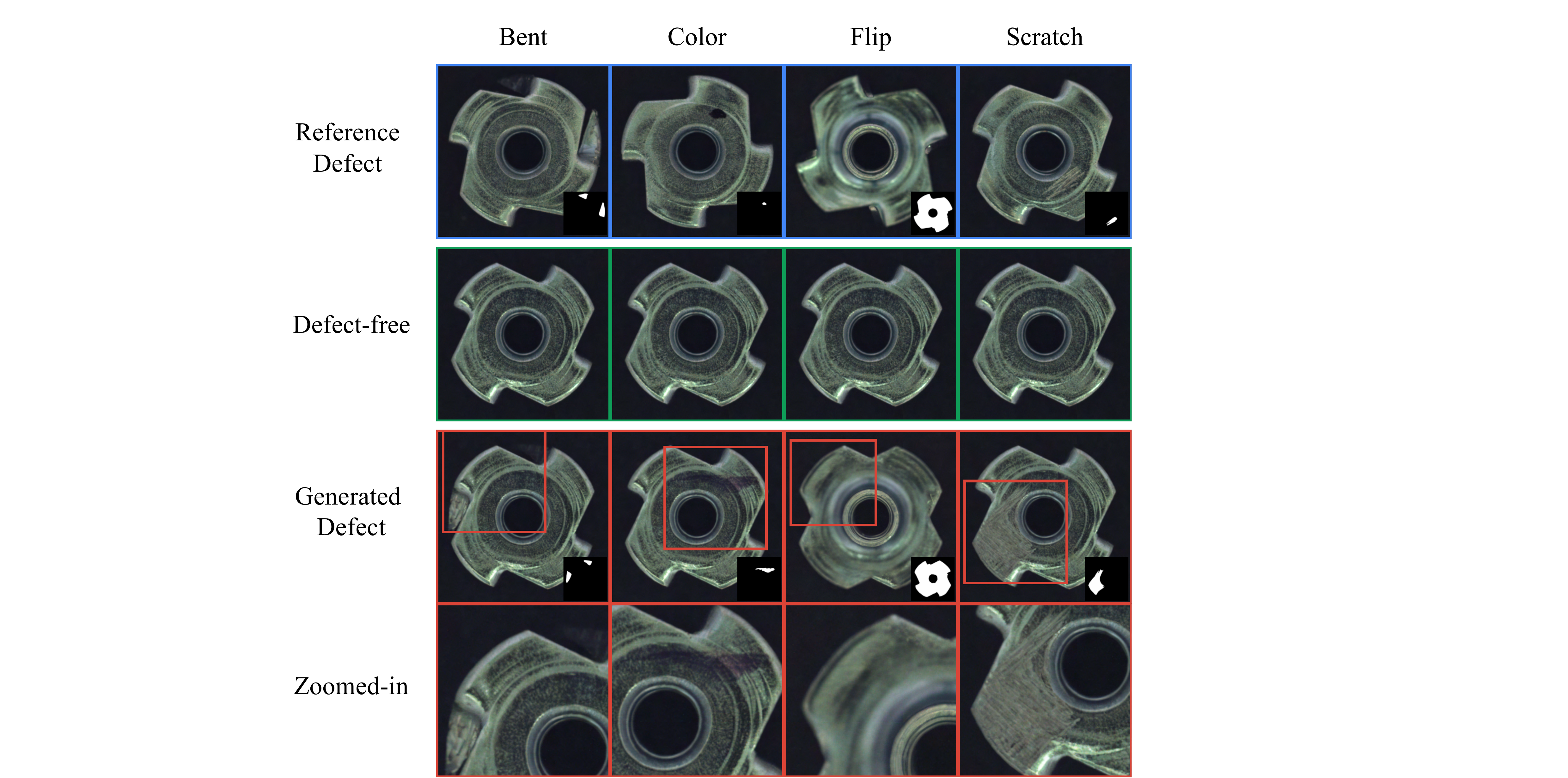}
    \caption{\textbf{Defect generation results on MVTec AD dataset (object: metal nut).}}
    \label{fig:metal_nut}
  \hfill
\end{figure*}

\begin{figure*}
  \centering
    \includegraphics[width=\textwidth]{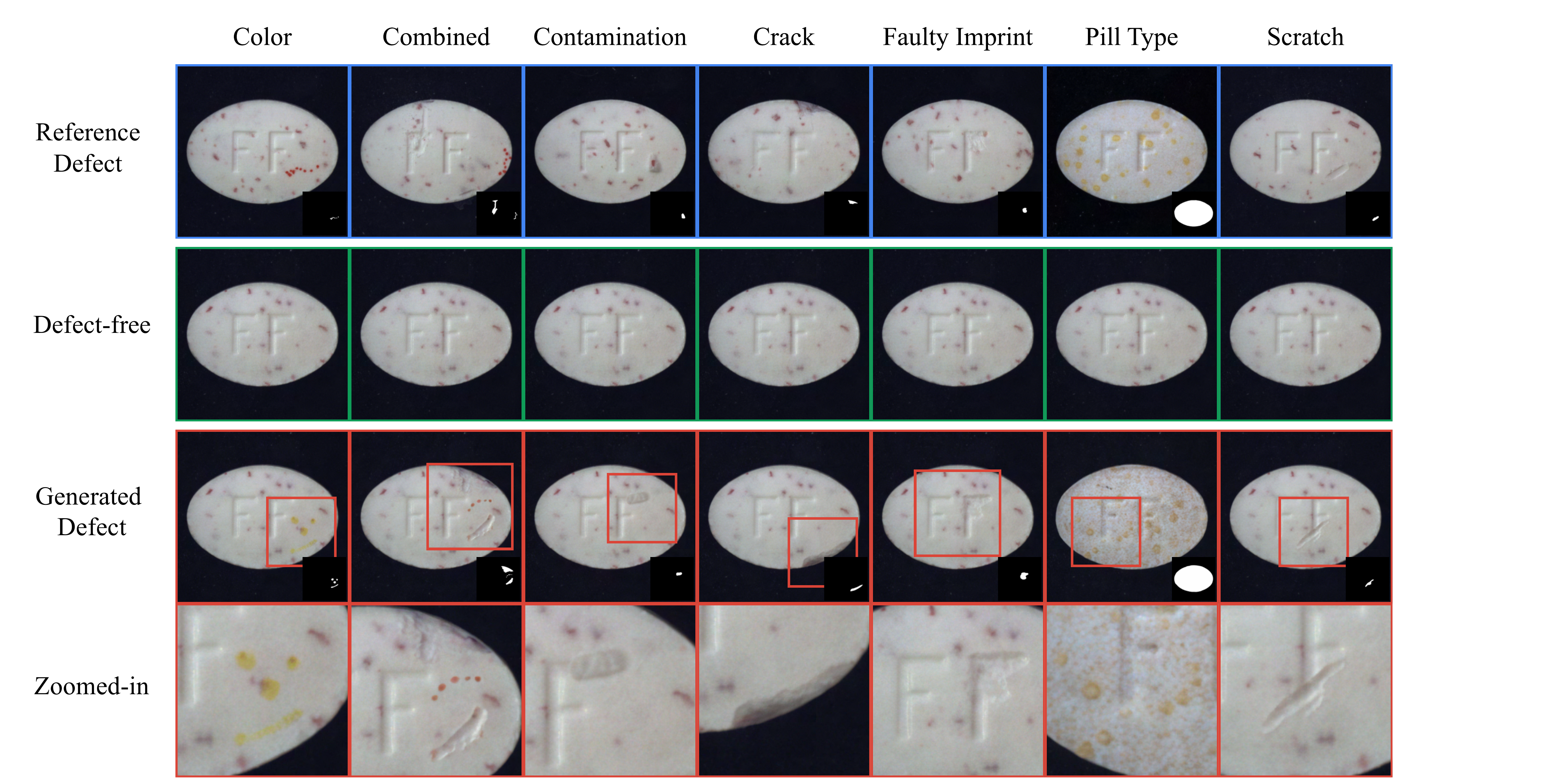}
    \caption{\textbf{Defect generation results on MVTec AD dataset (object: pill).}}
    \label{fig:pill}
  \hfill
\end{figure*}

\begin{figure*}
  \centering
    \includegraphics[width=\textwidth]{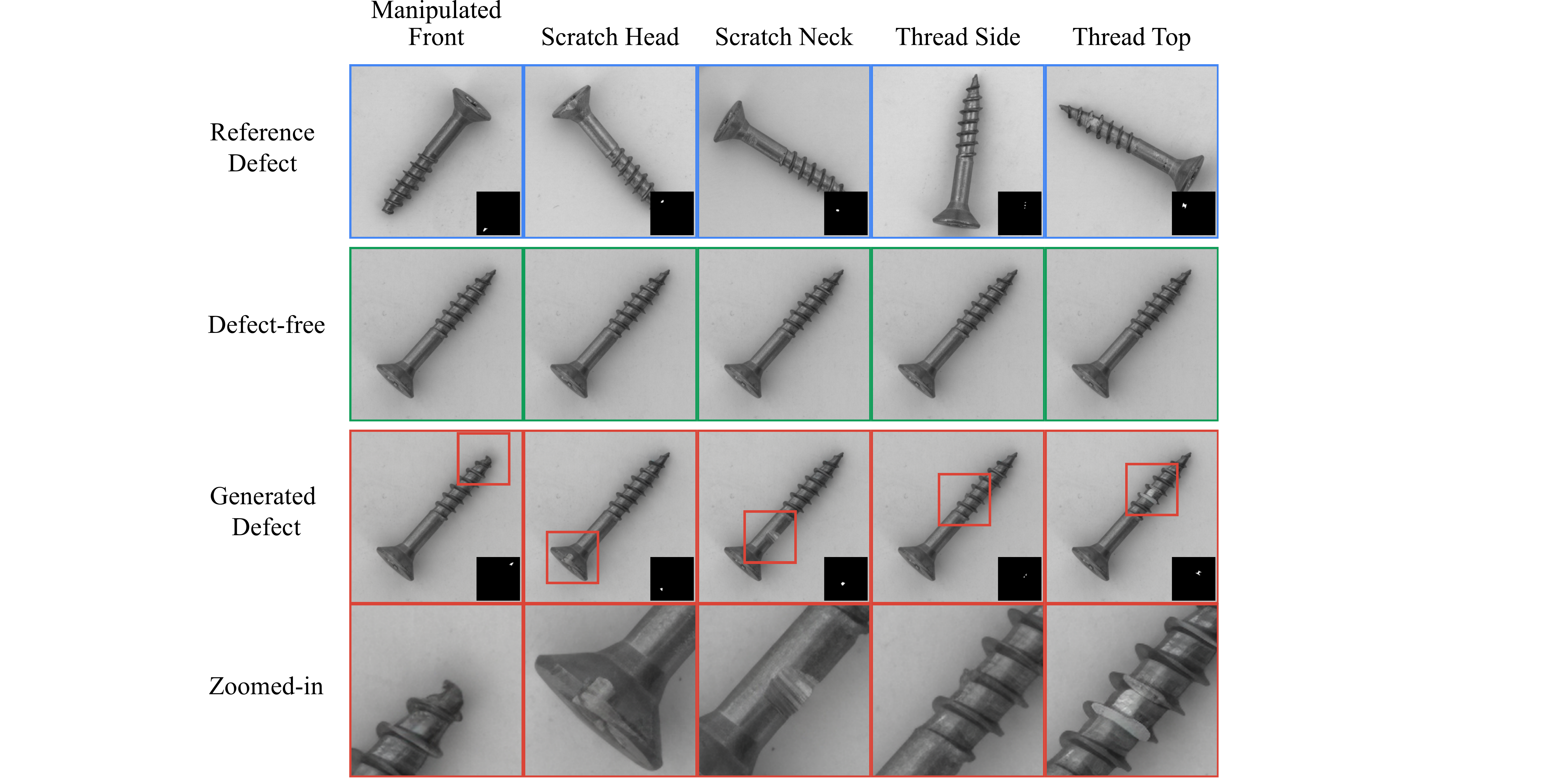}
    \caption{\textbf{Defect generation results on MVTec AD dataset (object: screw).}}
    \label{fig:screw}
  \hfill
\end{figure*}

\begin{figure*}
  \centering
    \includegraphics[width=\textwidth]{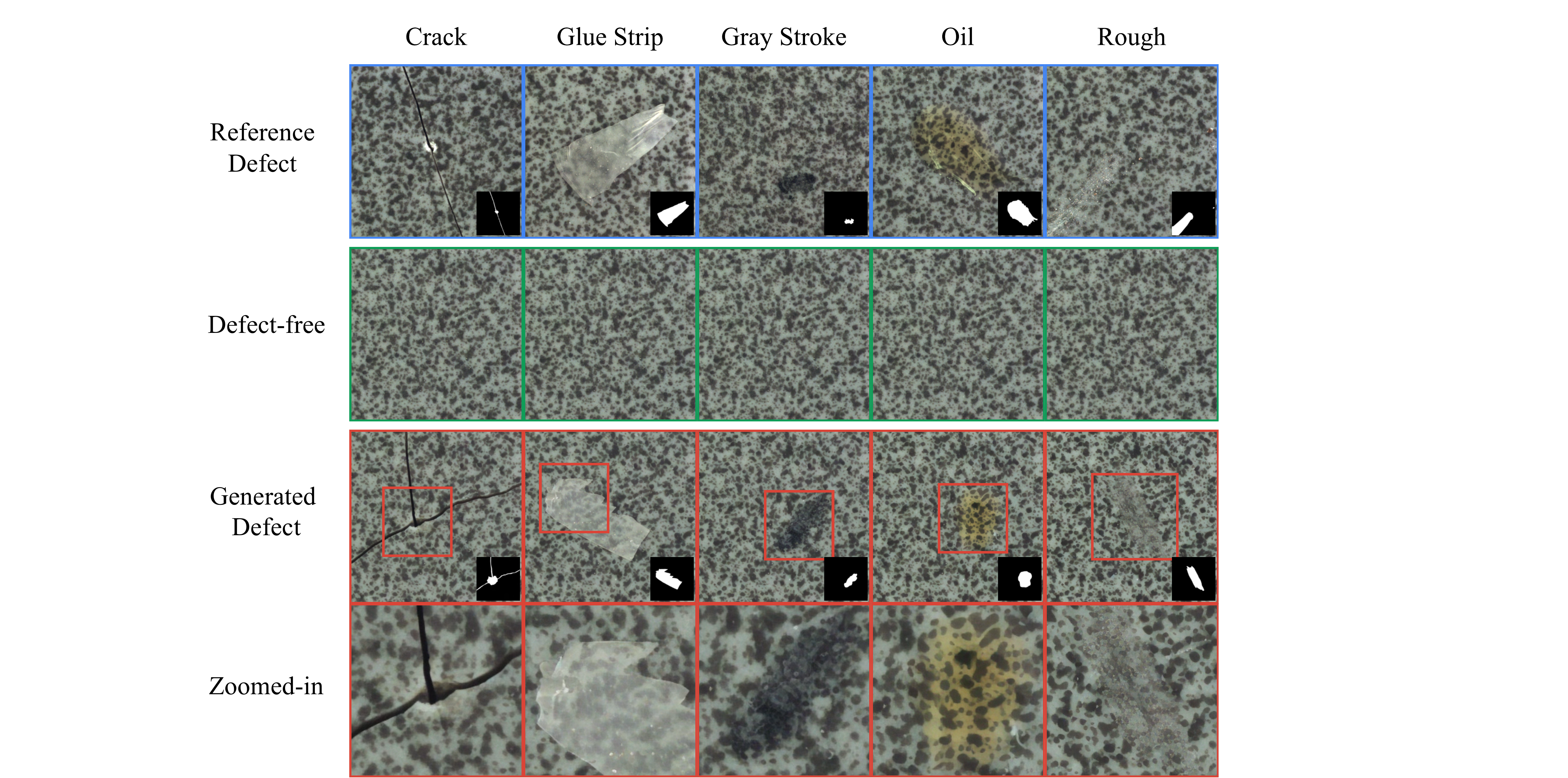}
    \caption{\textbf{Defect generation results on MVTec AD dataset (object: tile).}}
    \label{fig:tile}
  \hfill
\end{figure*}

\begin{figure*}
  \centering
    \includegraphics[width=\textwidth]{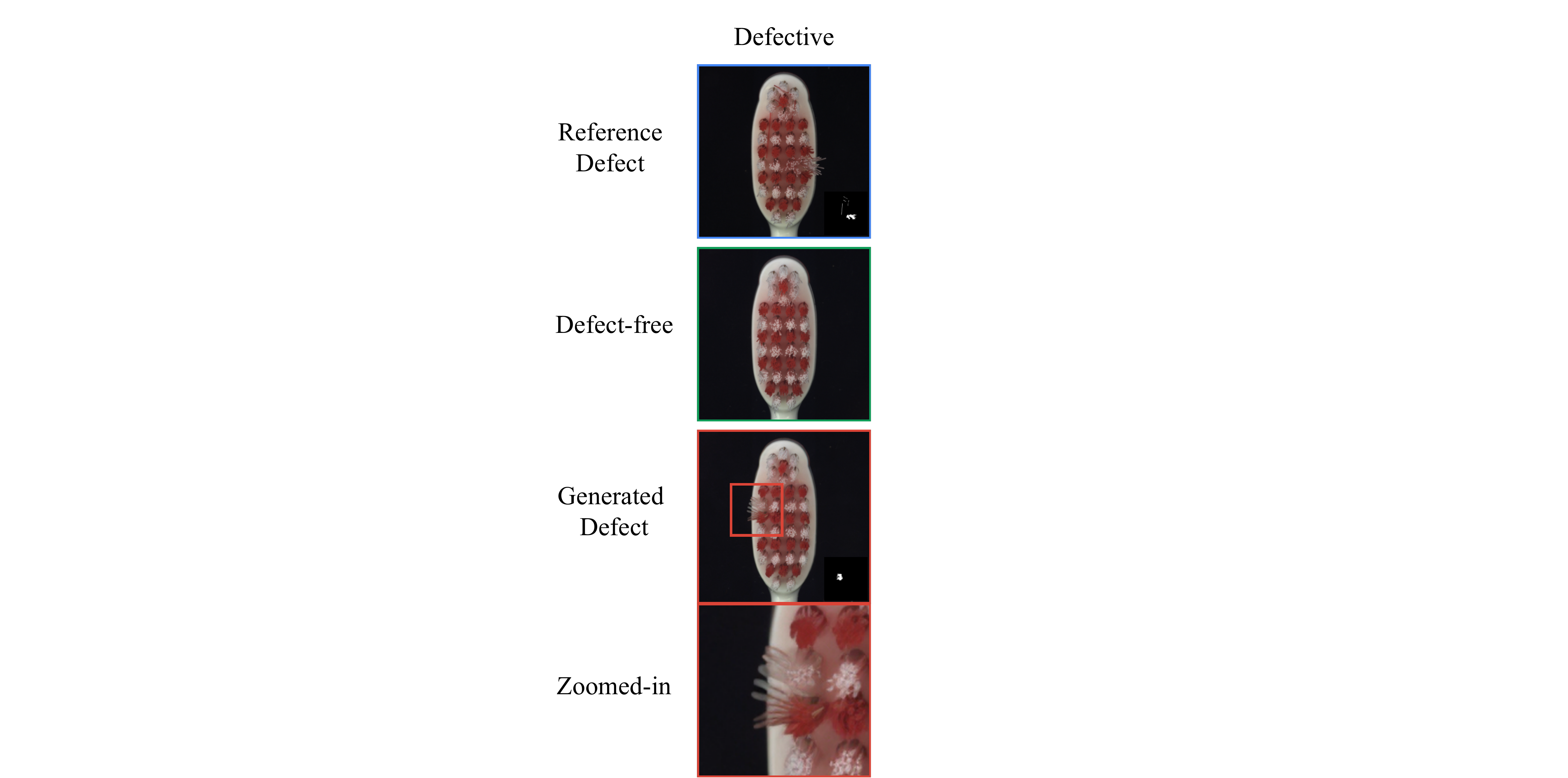}
    \caption{\textbf{Defect generation results on MVTec AD dataset (object: toothbrush).}}
    \label{fig:toothbrush}
  \hfill
\end{figure*}

\begin{figure*}
  \centering
    \includegraphics[width=\textwidth]{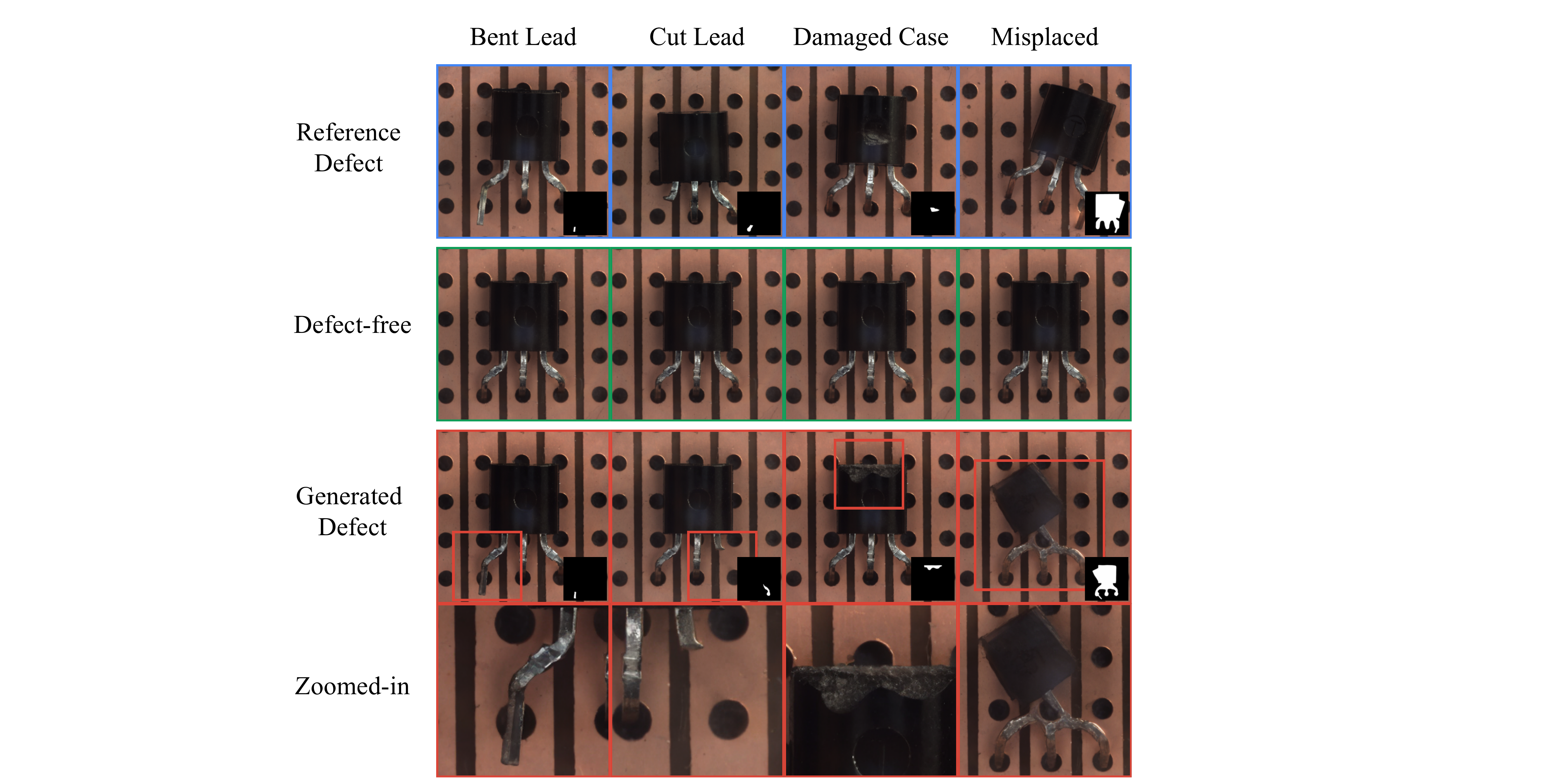}
    \caption{\textbf{Defect generation results on MVTec AD dataset (object: transistor).}}
    \label{fig:transistor}
  \hfill
\end{figure*}

\begin{figure*}
  \centering
    \includegraphics[width=\textwidth]{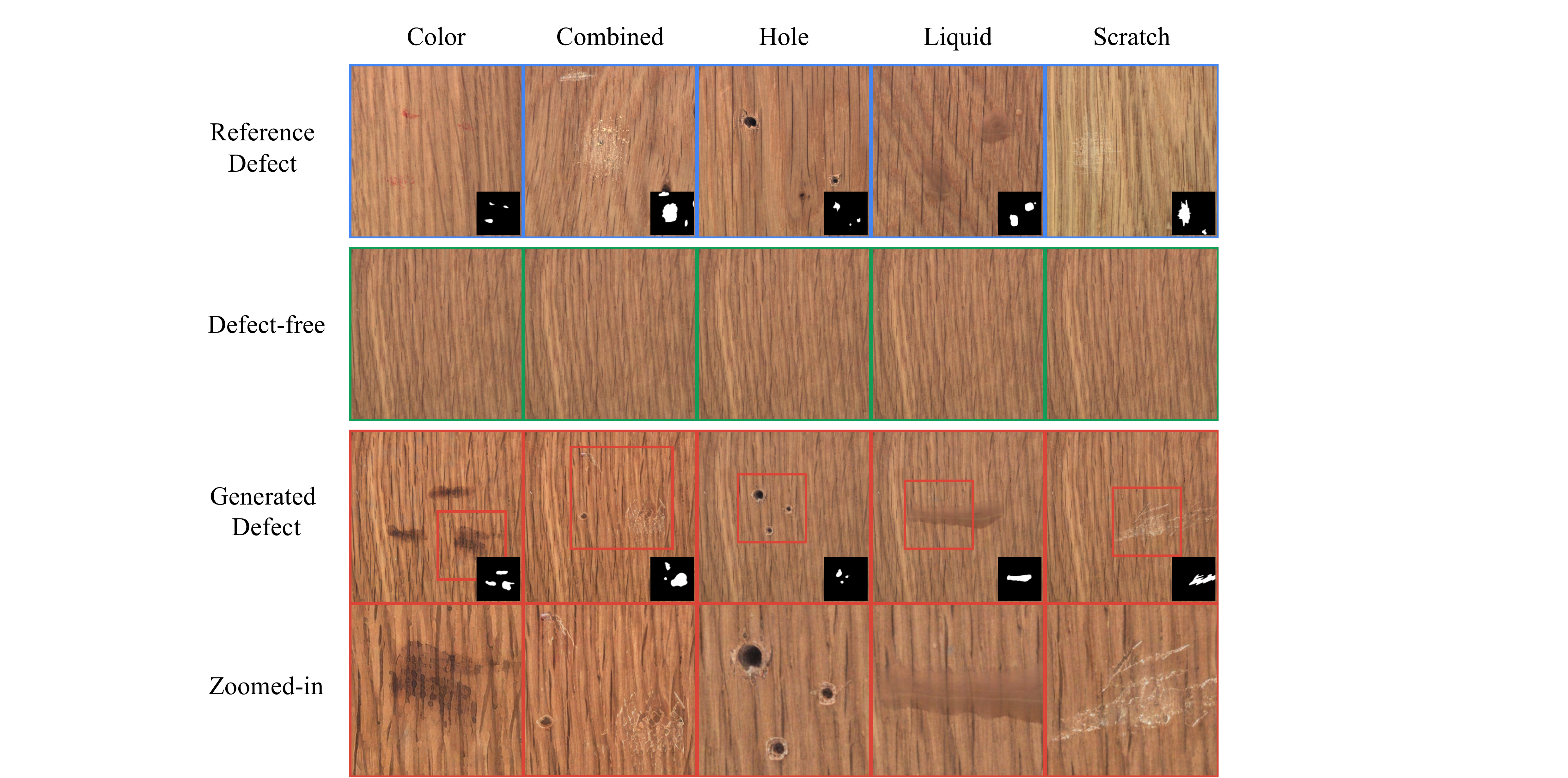}
    \caption{\textbf{Defect generation results on MVTec AD dataset (object: wood).}}
    \label{fig:wood}
  \hfill
\end{figure*}

\begin{figure*}
  \centering
    \includegraphics[width=\textwidth]{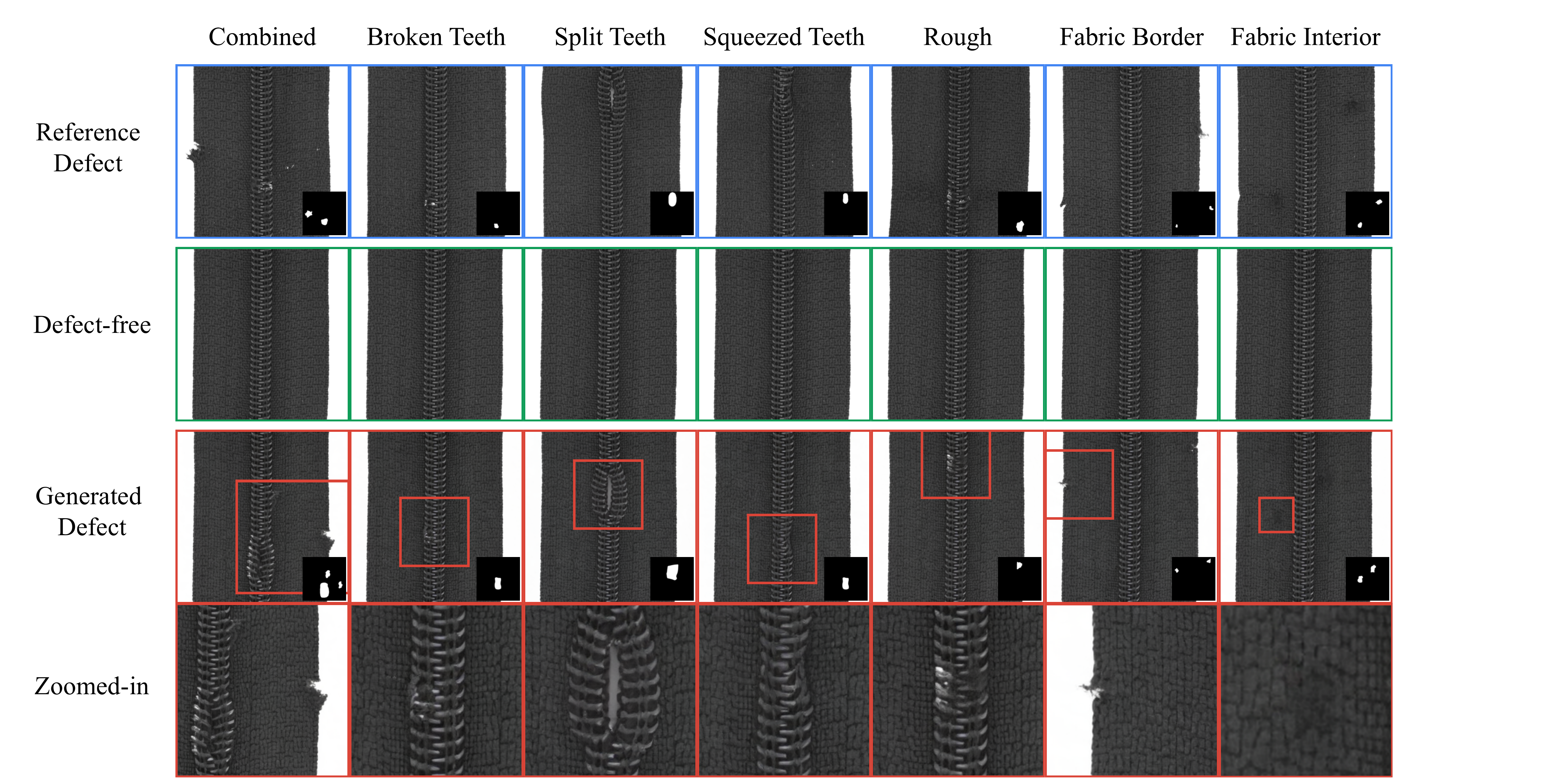}
    \caption{\textbf{Defect generation results on MVTec AD dataset (object: zipper).}}
    \label{fig:zipper}
  \hfill
\end{figure*}

We provide defect generation samples for all object and defect categories in the MVTec AD~\cite{bergmann2019mvtec} dataset. As illustrated in~\cref{fig:bottle,fig:cable,fig:capsule,fig:carpet,fig:grid,fig:hazelnut,fig:leather,fig:metal_nut,fig:pill,fig:screw,fig:tile,fig:toothbrush,fig:transistor,fig:wood,fig:zipper}, our method consistently generates realistic and naturally filled defects across all cases. The first row (blue box) displays the real defect images, while the second row (green box) contains the defect-free images used for defect generation. The third row presents the generated defects using the masks shown in the bottom-right corner, and the fourth row (red box) provides a zoomed-in view of the generated defects.

\subsection{VisA Dataset}

\begin{figure*}
  \centering
    \includegraphics[width=\textwidth]{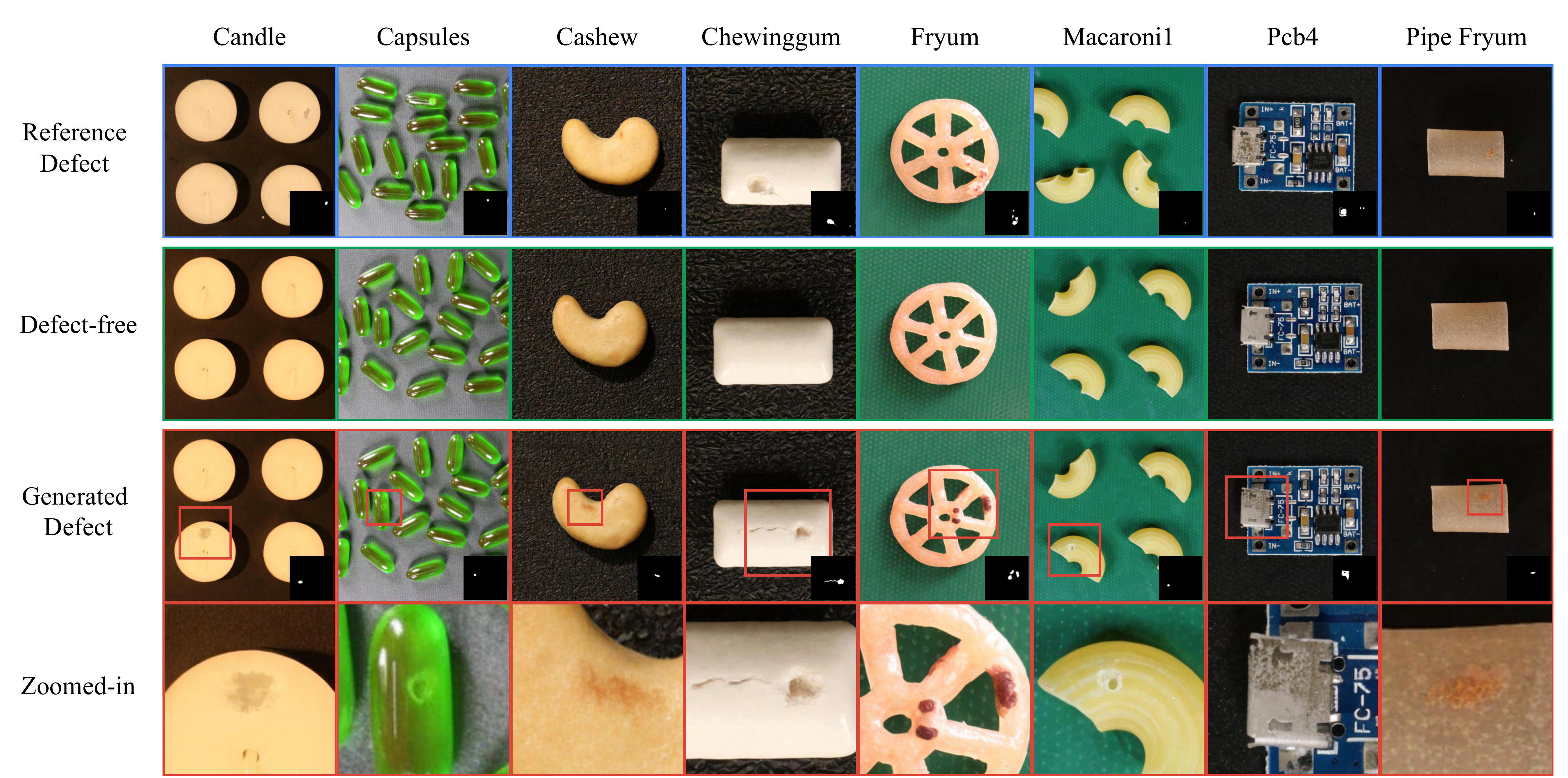}
    \caption{\textbf{Defect generation results on VisA dataset.}}
    \label{fig:VisA}
  \hfill
\end{figure*}

We further apply our method to another anomaly detection dataset, the Visual Anomaly (VisA)~\cite{zou2022spot} dataset. Following a similar approach to its application on MVTec AD dataset, we train the model using pairs of anomalous images and their corresponding masks (limited to the first 10 pairs per object) and generate defects on defect-free images using unseen masks. As shown in~\cref{fig:VisA}, our method successfully generates realistic defects across all object categories. This highlights the robustness of our method in generalizing to a variety of real-world defects.

\setlength{\tabcolsep}{6pt} 

\begin{table}[!htp]
    \centering
    \begin{adjustbox}{max width = 1.0\linewidth} 

    \begin{tabular}{c|cc|ccc}\toprule
    \multirow{2}{*}{Objects} &\multicolumn{2}{c}{Ours w/o LFS} &\multicolumn{2}{c}{Ours} \\\cmidrule{2-5}
        &KID↓ &IC-LPIPS↑ &KID↓ &IC-LPIPS↑ \\\midrule
        bottle &33.57 &\textbf{0.12} &\textbf{30.99} &\textbf{0.12} \\
        capsule &\textbf{5.01} &0.17 &5.60 &\textbf{0.18} \\
        carpet &50.39 &0.21 &\textbf{50.37} &\textbf{0.22} \\
        hazelnut &1.86 &\textbf{0.31} &\textbf{1.13} &\textbf{0.31} \\
        leather &83.06 &0.29 &\textbf{74.66} &\textbf{0.30} \\
        pill &16.22 &0.22 &\textbf{8.76} &\textbf{0.23} \\
        tile &49.59 &\textbf{0.44} &\textbf{45.14} &\textbf{0.44} \\
        toothbrush &\textbf{2.87} &\textbf{0.15} &3.19 &\textbf{0.15} \\
        wood &7.05 &\textbf{0.35} &\textbf{4.72} &\textbf{0.35} \\
        zipper &35.23 &\textbf{0.21} &\textbf{34.91} &0.20 \\
        \bottomrule
    \end{tabular}
    \end{adjustbox}
    
    \caption{\textbf{Generation Comparison with Low-Fidelity Selection.} The application of LFS demonstrates improvements in both quality (KID) and diversity (IC-LPIPS). The values represent averages calculated for each defect category.}
    \label{tab:lfs}
\end{table}

\begin{figure}[!tb]
  \centering
  \includegraphics[width=\columnwidth]{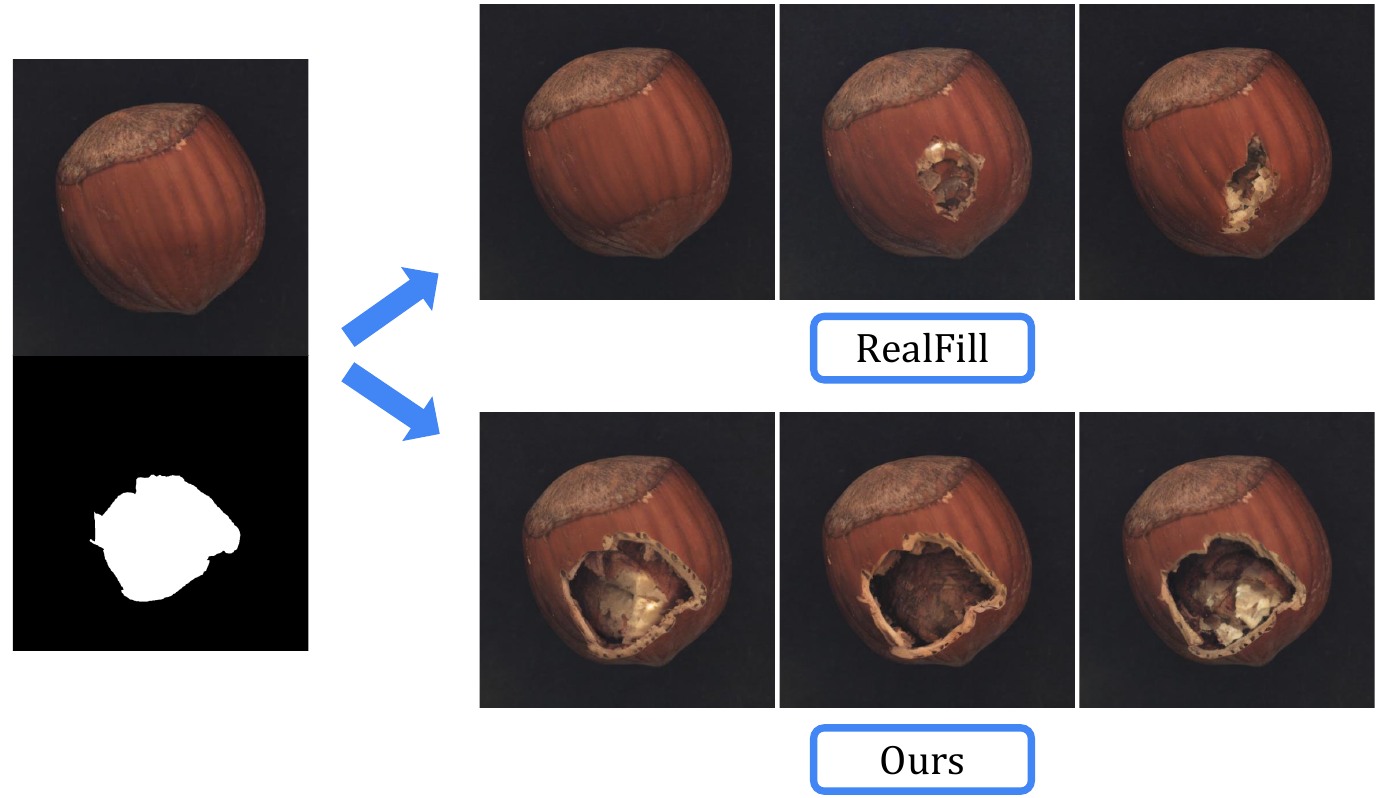}
  \caption{\textbf{Comparison to RealFill.} This figure shows a comparison of defect generation quality with another inpainting-based concept learning method, RealFill~\cite{tang2024realfill}. It fails to generate proper defects, either reconstructing the original region or producing unrealistic defects that are misaligned with the mask (upper images). In contrast, DefectFill (ours) generates realistic and diverse defects that align accurately with the mask (lower images).}
  \label{fig:realfill comparison}
\end{figure}

\section{Additional Quantitative Results}
\label{sec:additional quantitative results}

\subsection{Low Fidelity Selection}
\label{subsec:lfs}

\begin{table*}[!htb]
    \centering

    \begin{adjustbox}{max width = 1.0\linewidth}
    \begin{tabular}{c|ccc|ccc|ccc|ccc}
        \toprule
        \multirow{2}{*}{Objects} &\multicolumn{3}{c}{DFMGAN†} &\multicolumn{3}{c}{AnoDiff*} &\multicolumn{3}{c}{AnoDiff‡} &\multicolumn{3}{c}{Ours} \\\cmidrule{2-13}
        &AUROC↑ &AP↑ &$F_1$-max↑ &AUROC↑ &AP↑ &$F_1$-max↑ &AUROC↑ &AP↑ &$F_1$-max↑ &AUROC↑ &AP↑ &$F_1$ max↑ \\\midrule
        bottle &0.97 &\textbf{1.00} &0.98 &\textbf{1.00} &\textbf{1.00} &0.99 &\textbf{1.00} &\textbf{1.00} &\textbf{1.00} &\textbf{1.00} &\textbf{1.00} &\textbf{1.00} \\
        capsule &0.76 &0.9 &0.87 &\textbf{1.00} &\textbf{1.00} &\textbf{0.99} &0.94 &0.98 &0.93 &0.98 &\textbf{1.00} &0.97 \\
        carpet &0.81 &0.92 &0.82 &0.97 &0.99 &0.94 &0.89 &0.95 &0.88 &\textbf{1.00} &\textbf{1.00} &\textbf{1.00} \\
        hazelnut &\textbf{1.00} &\textbf{1.00} &0.99 &\textbf{1.00} &\textbf{1.00} &0.99 &\textbf{1.00} &\textbf{1.00} &0.98 &\textbf{1.00} &\textbf{1.00} &\textbf{1.00} \\
        leather &0.94 &0.97 &0.92 &\textbf{1.00} &\textbf{1.00} &\textbf{1.00} &\textbf{1.00} &\textbf{1.00} &\textbf{1.00} &\textbf{1.00} &\textbf{1.00} &\textbf{1.00} \\
        pill &0.92 &0.97 &0.92 &\textbf{0.98} &\textbf{1.00} &\textbf{0.97} &0.97 &0.99 &0.95 &0.97 &0.99 &0.95 \\
        tile &\textbf{1.00} &\textbf{1.00} &0.99 &\textbf{1.00} &\textbf{1.00} &\textbf{1.00} &\textbf{1.00} &\textbf{1.00} &\textbf{1.00} &\textbf{1.00} &\textbf{1.00} &\textbf{1.00} \\
        toothbrush &0.97 &0.98 &0.93 &\textbf{1.00} &\textbf{1.00} &\textbf{1.00} &\textbf{1.00} &\textbf{1.00} &\textbf{1.00} &\textbf{1.00} &\textbf{1.00} &0.98 \\
        wood &0.89 &0.94 &0.87 &0.98 &0.99 &0.99 &0.98 &0.99 &0.98 &\textbf{1.00} &\textbf{1.00} &\textbf{1.00} \\
        zipper &0.99 &\textbf{1.00} &\textbf{0.99} &\textbf{1.00} &\textbf{1.00} &\textbf{0.99} &\textbf{1.00} &\textbf{1.00} &\textbf{0.99} &\textbf{1.00} &\textbf{1.00} &\textbf{0.99} \\
        \bottomrule
    \end{tabular}
    \end{adjustbox}

    \caption{\textbf{Image-Level Detection Comparison.} The table presents AUROC, AP, and $F_1$-max scores for image-level anomaly detection evaluation using a UNet trained on generated defect images. Our method achieves the highest performance across most metrics and objects. The labels are defined in Tab. 2.
    }
    \label{tab:detection}
    
\end{table*}
\begin{table}[!tp]\centering
    \scriptsize
    \resizebox{\columnwidth}{!}{ 
    \begin{tabular}{lrrrrrrrr}\toprule
\multicolumn{2}{c}{w/o $\mathcal{L}_{obj}$} &\multicolumn{2}{c}{w/o $\mathcal{L}_{def}$} &\multicolumn{2}{c}{w/o $\mathcal{L}_{attn}$} &\multicolumn{2}{c}{Ours} \\\cmidrule{1-8}
KID↓ &IC-LPIPS↑ &KID↓ &IC-LPIPS↑ &KID↓ &IC-LPIPS↑ &KID↓ &IC-LPIPS↑ \\\midrule
26.26 &\textbf{0.25} &67.34 &0.23 &26.51 &0.24 &\textbf{25.95} &\textbf{0.25} \\
\bottomrule
\end{tabular}
}
\caption{Results when each loss term is removed during training.}
\label{tab:loss_ablation}
\end{table}

\cref{tab:lfs} compares the quality (KID~\cite{binkowski2018demystifying}) and diversity (IC-LPIPS~\cite{ojha2021few}) of generated defect images with and without applying Low-Fidelity Selection (LFS). For diversity, applying LFS achieves the best performance across all objects except for the zipper. In terms of quality, applying LFS improves the KID score for all objects except the capsule and toothbrush.

\subsection{Detection}
\label{subsec:detection}

Similar to the evaluation of the anomaly localization task (Tab. 2), we also evaluate our method on the image-level anomaly detection task, comparing it with defect generation baselines (DFMGAN~\cite{duan2023few}, AnomalyDiffusion~\cite{hu2024anomalydiffusion}). \cref{tab:detection} shows our method achieves the best scores in most cases. Even in instances where it does not achieve the best score, it consistently performs well, with all scores exceeding 0.95.

\subsection{Loss Ablation}
\label{subsec:loss_ablation}
\cref{tab:loss_ablation} shows the evaluation results on the MVTec dataset after removing each loss term during training. Notably, removing $\mathcal{L}_{def}$ causes a significant increase in KID. Using all terms achieves the best scores for both KID and IC-LPIPS.

\section{Comparison to RealFill}
\label{sec:inpainting concept learning comparison}

To demonstrate DefectFill's ability to learn defect features and generate realistic defects, we compare it with another inpainting-based concept learning method, RealFill~\cite{tang2024realfill}. While RealFill focuses on filling erased regions in a single target image, making it less suitable for defect generation tasks required in visual inspection, this comparison highlights the superior generation quality of DefectFill.
As shown in~\cref{fig:realfill comparison}, RealFill (upper images) fails to generate proper defects, often reconstructing the original region or producing unrealistic defects that are misaligned with the mask. In contrast, our method (lower images) generates defects that are both realistic and diverse, while precisely aligning with the mask's shape. This highlights not only the importance of leveraging an inpainting diffusion model but also the crucial role of our defect-specific loss, which is tailored for inpainting diffusion models.

\begin{figure}[t!]
  \centering
  \includegraphics[width=\columnwidth]{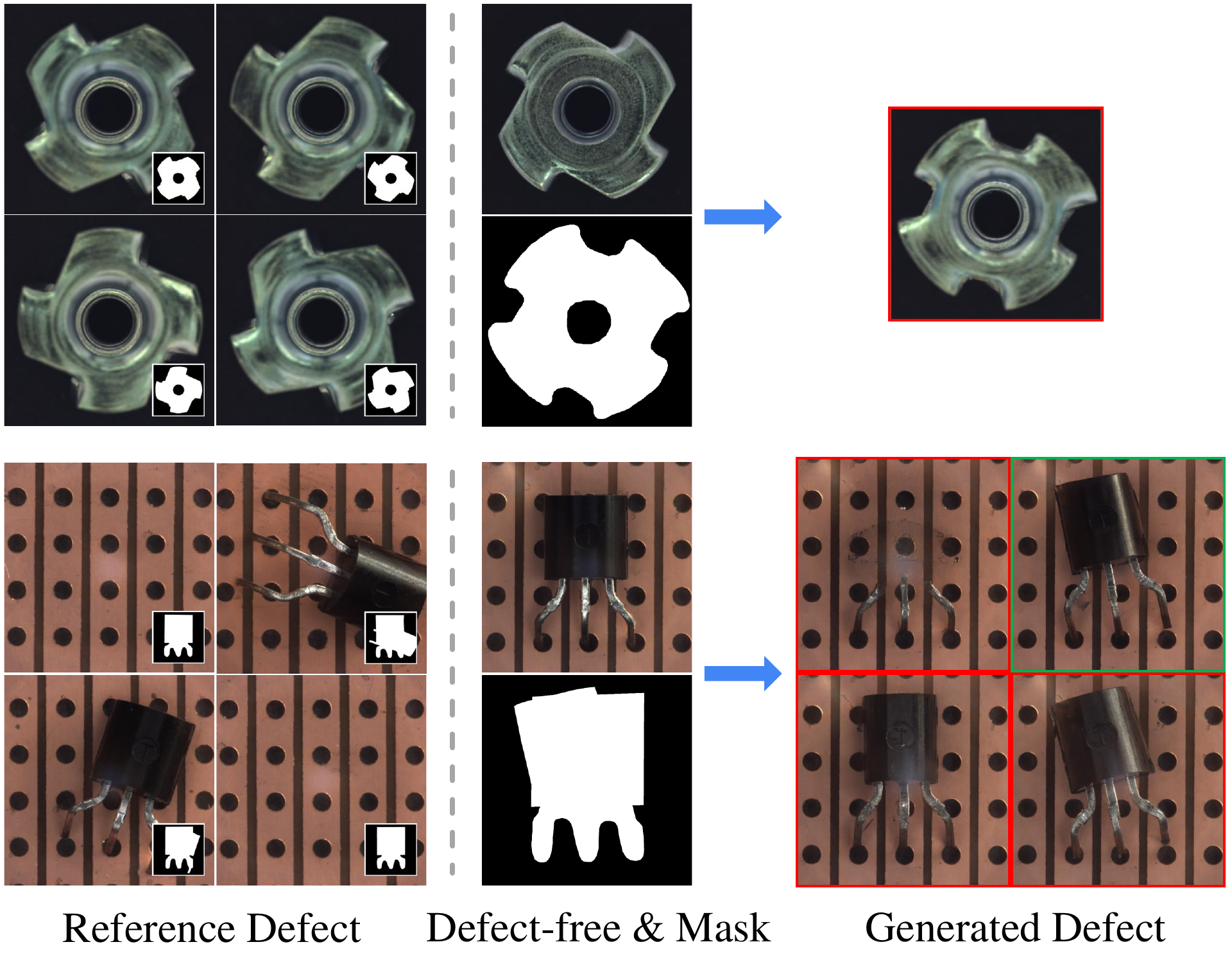}
  \caption{\textbf{Failure Cases.} DefectFill struggles with structural defects affecting the entire object. For the metal nut (top), the mask covers the flipped nut itself, so the model learns its appearance rather than its orientation. For the transistor (bottom), inpainting replaces the defect-free object, creating a stochastic mix of defect features, though it often generates proper defects (green box).}
  \label{fig:failure}
\end{figure}

\section{Failure Cases}
\label{sec:failure cases}

As discussed in the conclusion, our method excels at generating local defects but is less effective at handling global structural defects.
~\cref{fig:failure} illustrates failure cases of structural defects from the MVTec AD dataset.
For the metal nut's \textit{flip} defect (upper part of~\cref{fig:failure}), both the reference defect image and mask represent the entire flipped nut. This causes the model to learn the flipped nut's appearance rather than the direction of the flipped teeth as a defect feature. Consequently, when generating a flipped nut from an unflipped one, the teeth's direction remains unchanged, and the model instead fills the appearance aligning with the mask shape.
For the transistor's \textit{misplaced} defect (lower part of~\cref{fig:failure}), the scenario differs. The mask includes both the original and misaligned positions, enabling the model to learn misalignment features. However, the \textit{misplaced} defect involves not only misaligned cases but also missing ones. In this situation, the inpainting process entirely removes the transistor from the original position and generates a new defect. This results in the loss of semantic information from the defect-free object, causing stochastic appearances of defect features representing both misaligned and missing cases. As shown in~\cref{fig:failure}, the generated defects manifest as complete transparency, semi-transparent alignment, semi-transparent misalignment (red boxes), or proper misalignment (green box).
Addressing these global structural defects is left for future research. Nevertheless, our method demonstrates strong performance in handling most practical cases, where localized defects are the primary focus in real-world scenarios.

\begin{figure}
  \centering
  \includegraphics[width=\columnwidth]{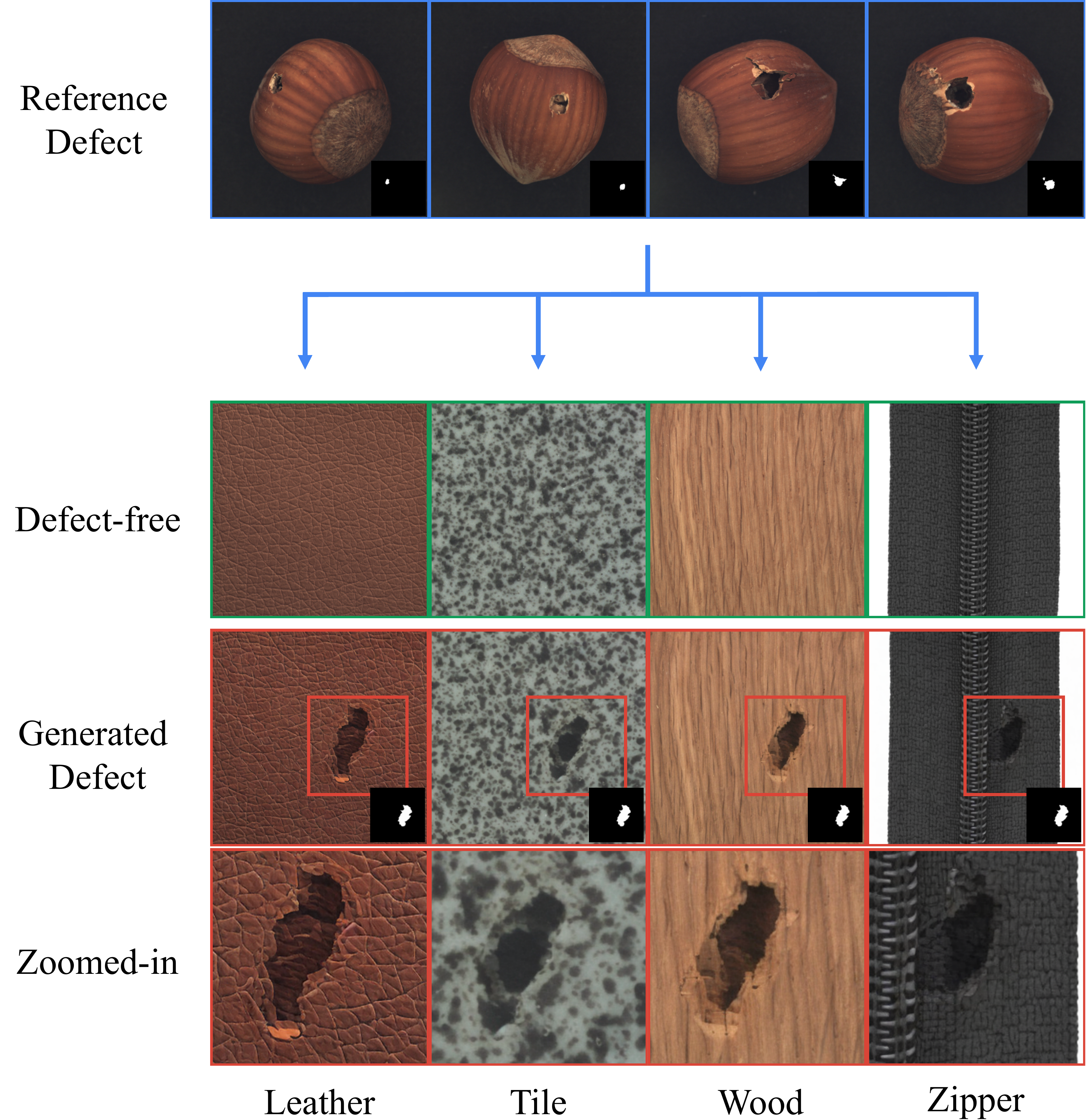}
  \caption{\textbf{Transferring defects across different objects.} The figure illustrates the results of generating hole defects in different objects after learning the features of a hole defect from a hazelnut. Defect transfer can occur when the defect features are general and plausible in the context of other objects.}
  \label{fig:object transfer}
\end{figure}

\section{Transferring Defects across Objects}
\label{sec:defect transfer}

We observe that if a defect in one object exhibits general features, it can be generated in other objects where such a defect is plausible.
As shown in~\cref{fig:object transfer}, after learning the hole defect from a hazelnut, our method successfully generates similar defects in various defect-free objects (\eg leather, zipper, wood, and tile).


\end{document}